\documentclass[sigconf]{acmart}

\newcommand{\harada}[1]{\textcolor{blue}{(Harada: {#1})}}

\def\0{\mathbf{0}}

\usepackage[utf8]{inputenc}

\usepackage{soul}
\usepackage[utf8]{inputenc}
\usepackage{amsmath}
\usepackage{amsfonts}
\usepackage{booktabs}
\usepackage[ruled]{algorithm2e} %
\usepackage{color, colortbl}

\usepackage{subcaption,siunitx,booktabs}
\usepackage{multirow}
\usepackage{bm}
\usepackage[normalem]{ulem}
\usepackage{balance}

\newtheorem{theorem}{Theorem}

\usepackage{amsthm}
\theoremstyle{definition}

\newtheorem{assumption}{Assumption}

\definecolor{Gray}{gray}{0.95}

\usepackage[first=0,last=9]{lcg}

\AtBeginDocument{%
  \providecommand\BibTeX{{%
    \normalfont B\kern-0.5em{\scshape i\kern-0.25em b}\kern-0.8em\TeX}}}

\copyrightyear{2021}
\acmYear{2021}
\setcopyright{rightsretained}
\acmConference[CIKM '21]{Proceedings of the 30th ACM International Conference on Information and Knowledge Management}{November 1--5, 2021}{Virtual Event, QLD, Australia}
\acmBooktitle{Proceedings of the 30th ACM International Conference on Information and Knowledge Management (CIKM '21), November 1--5, 2021, Virtual Event, QLD, Australia}
\acmDOI{10.1145/3459637.3482349}
\acmISBN{978-1-4503-8446-9/21/11}

\ccsdesc[500]{Computing methodologies~Machine learning approaches}

\begin{document}
\fancyhead{}

\title[GraphITE]{GraphITE: Estimating Individual Effects of \\ Graph-structured Treatments}

\author{Shonosuke Harada}
\email{sh1108@ml.ist.i.kyoto-u.ac.jp }
\affiliation{ \institution{Kyoto University}   }

\author{Hisashi Kashima}
\email{kashima@i.kyoto-u.ac.jp}
\affiliation{ \institution{Kyoto University}   }

\renewcommand{\shortauthors}{Shonosuke Harada and Hisashi Kashima}

\sloppy

\begin{abstract}
 Outcome estimation of treatments for individual targets is a crucial foundation for decision making based on causal relations.
Most of the existing outcome estimation methods deal with binary or multiple-choice treatments; 
however, in some applications, the number of interventions can be very large, while the treatments themselves have rich information.
In this study, we consider one important instance of such cases, that is, the outcome estimation problem of graph-structured treatments such as drugs.
Due to the large number of possible interventions, the counterfactual nature of observational data, which appears in conventional treatment effect estimation, becomes a more serious issue in this problem.
Our proposed method GraphITE (pronounced `graphite') obtains the representations of the graph-structured treatments using graph neural networks, and also mitigates the observation biases by using the HSIC regularization that increases the independence of the representations of the targets and the treatments. 
In contrast with the existing methods, which cannot deal with ``zero-shot" treatments that are not included in observational data, GraphITE can efficiently handle them thanks to its capability of incorporating graph-structured treatments. 
The experiments using the two real-world datasets show GraphITE outperforms baselines especially in cases with a large number of treatments.

\end{abstract}

\keywords{causal inference, treatment effect estimation, graph neural networks}

\maketitle

 \begin{figure}[tb]
\centering
  \begin{center}
\includegraphics[width=1.02\linewidth]{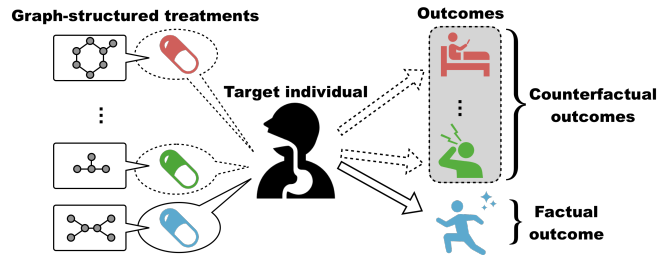}
      \centering
\end{center}
 \caption{\small{Individual effect estimation problem of graph-structured treatments. The possible treatments, i.e., drugs, are associated with graphs representing their molecular structures. 
 In observational data, only one treatment is applied to the target individual; consequently, only the factual outcome is observed, while the counterfactual outcomes for the other treatments are not. Our goal is to predict the outcomes of all treatments for future targets.}}\label{fig:intro}
\end{figure}

\section{Introduction}
Estimating causal effects of treatments for individual targets, which is often referred to as individual treatment effect~(ITE), is an important foundation for efficient decision making based on observational data. 
The scope of applications of causal inference ranges across a wide range of fields, including medicine, education, and economic policy. 
The main difficulties in estimating ITE are (i) the counterfactual nature of observational data, that is, only the outcome of an actual treatment is observed and (ii) biases in observational data due to biases in past treatment decisions. 
To address these difficulties, various statistical techniques have been developed, including matching~\cite{rubin1973matching}, inverse propensity score weighting~\cite{rosenbaum1983central}, instrumental variable methods~\cite{baiocchi2014instrumental}, as well as more modern representation learning approaches~\cite{shalit2017estimating,johansson2016learning}.

Most previous studies dealt with a binary or relatively small number of treatments. 
However, in some scenarios, the number of treatments can be considerably larger. 
For example, when modeling drug effects on target cells, the number of candidate drugs (i.e., treatments) can be huge, while the number of observations per drug can be quite small due to the high cost of clinical trials~\cite{ganter2005development,hoelder2012discovery}.
Each drug is composed of a bunch of atoms, such as carbon, oxygen, and nitrogen, and the number of drugs composed by the combination of atoms are substantially huge.
A similar situation can also occur in online advertisements~\cite{saini2019multiple}. 
This scarcity of data exacerbates the aforementioned problems. 
More seriously, some treatments that never appeared in the observational data, such as new drugs or new ads, may appear for the first time during the test phase. 
Despite the significant importance of estimating unobserved treatment effects in various applications, existing methods are not capable of dealing with such ``zero-shot" treatment effect estimation.

In this study, we consider the individual treatment effect estimation problem with a large number of treatments, for which there is no definitive existing solution due to extremely sparse observations.
To solve the problem, we focus on auxiliary information that accompanies the treatments.
Such auxiliary information is sometimes available in applications.
In the drug effect example, each drug is a chemical compound with its own molecular structure, which can be represented as a graph (Figure~\ref{fig:intro}), and it is expected to take advantage of structural patterns contained in the graph structure.
The rich structural information of graphs allows us to transfer useful information for predicting outcomes from treatments with many observations to those with less observations, even to ``zero-shot" treatments that have not been seen before.
Therefore, our challenge in this paper is to incorporate the rich graph structure information of treatments into our treatment effect estimation model, and provide an effective learning method to mitigate biases in sparse observational data.

We propose GraphITE (pronounced ``graphite''), which is an outcome prediction model for graph-structured treatments based on biased observational data.
It is built upon the recent significant advances in learning representations using graph neural networks~(GNNs)~\cite{kipf2016semi,gilmer2017neural}.
Bias mitigation with a large (possibly infinite) number of treatments is another issue because most existing frameworks~\cite{shalit2017estimating,schwab2018perfect,saini2019multiple} are not designed for such cases.
To reduce the treatment selection bias depending on the individual target, GraphITE finds representations of the target and treatment that are as independent of each other as possible. This is achieved by Hilbert-Schmidt Independence Criterion~(HSIC)~regularization, which was recently proposed by Lopez et al.~(\citeyear{lopez2020cost});
we extend their framework  to exploit the treatment features extracted by GNNs and give theoretical justification on how reducing biases over the representation space extracted from graph space leads to unbiased results.
Our formulation makes it possible to reduce the selection bias caused by complex graph structure information, even for the zero-shot treatments that cannot be handled by existing frameworks.

We conduct experiments on two real datasets: one with a relatively small number of treatments and one with over $100$ treatments.
The results show that the graph structures contribute to improving the predictive performance and that the HSIC regularization is robust to the presence of selection bias.

The contributions of this study can be summarized as follows:
\begin{itemize}
\item We propose GraphITE, an outcome prediction model for graph-structured treatments,
that can exploit auxiliary information of treatments to deal with a large number of treatments and ``zero-shot" treatments.
\item In order to train GraphITE from biased observational data, we extend HSIC regularization to cases where treatments have features, and give theoretical justification on how HSIC regularization making use of representations extracted from graph space contributes to mitigating biases.
\item Experiments on two real-world datasets empirically demonstrate the benefits of GraphITE for biased observational data and zero-shot treatments.
\end{itemize}
\section{Related Work\label{sec:related}}
\subsubsection*{Treatment effect estimation.}
Treatment effect estimation is a practically important task and has been widely studied in various fields ranging from  healthcare~\cite{eichler2016threshold} and economy~\cite{lalonde1986evaluating} to education~\cite{zhao2017estimating}. 

One of the typical solutions is the matching method~\cite{rubin1973matching,abadie2006large}, which compares the outcomes for pairs with similar covariates but different treatments. The propensity score, which is the probability of a target individual receiving a treatment, is introduced to mitigate the curse of dimensionality and selection bias~\cite{rosenbaum1983central}.
Tree-based methods, such as Causal Forest~\cite{wager2018estimation} and BART~\cite{chipman2010bart,hill2011bayesian}, have also been proposed and shown promising performances. 

Recently, representation learning based on deep neural networks has been successfully applied to treatment effect estimation and outperformed traditional methods~\cite{shalit2017estimating,johansson2016learning,yao2018representation}. 
They encourage the representations of treatment and control groups to get closer to each other to reduce selection bias. 
In addition, confounding variables are extracted by an additional neural network that predicts treatment assignments~\cite{NEURIPS2019_8fb5f8be}.
Generative adversarial neural networks~(GAN)~\cite{goodfellow2014generative} were also successfully applied to ITE estimation~\cite{bica2021real,yoon2018ganite}; their key idea is to train a predictive model (i.e., a generator) whose outcomes are difficult to distinguish between factual outcomes and counterfactual outcomes.

Most previous studies have focused on binary treatments, and extensions to multiple types of treatments, especially high numbers, are key research directions.
There have been several approaches designed for multiple treatments~\cite{schwab2018perfect,wager2018estimation,chipman2010bart}; however, most of them are limited to a relatively  small number of treatments, making it difficult to consider more than a few dozen treatments.
Saini et al.~\cite{saini2019multiple} whose motivation was somewhat similar to ours, considered combinatorial treatments; however, their focus was on a large number of combinations made from a small number of treatments, whereas we focus on many single treatments with the help of information on the treatments.

Extensions to real-valued treatments are also important for real-world applications, such as estimation of appropriate drug dosages~\cite{schwab2019learning,imbens2000role}.
Wang et al.~\cite{wang2020continuously} proposed an interesting approach to learn input representations that cannot distinguish real-valued domains.
Lopez et al.~\cite{lopez2020cost} considered 
total-ordered treatment spaces.
They proposed HSIC regularization for dealing with biased observational data; 
the theory of our proposed GraphITE is based on their theoretical framework, but  we extend the implications of their framework to representation learning of treatments with rich features.

\subsubsection*{Graph neural networks.}
Graph-structured data is one of the most popular data structures and has been widely employed in various domains such as social network analysis, citation analysis, and chemical informatics.
The GNN is one of the most successful deep neural network architectures owing to the practical importance of graph-structured data, and it has significantly improved the performance on various graph-structured data analysis tasks, such as node classification~\cite{kipf2016semi}, graph classification~\cite{duvenaud2015convolutional,gilmer2017neural}, and link prediction~\cite{DBLP:conf/nips/ZhangC18}, beyond conventional methods~\cite{hamilton2017inductive,henaff2015deep}.
In the field of chemo-informatics, GNNs have particularly flourished and played an important role in predicting molecule properties~\cite{duvenaud2015convolutional,schutt2018schnet,gilmer2017neural}, finding interactions between chemical objects~\cite{harada2020dual}, and generating desirable and unique molecules~\cite{you2018graph,zang2020moflow}.
 GraphITE also relies on their powerful  ability to extract features from graph-structured treatments.

Theoretical analysis of the expressive power of GNNs has been of great interest to researchers, for example, in their invertibility~\cite{xu2018powerful,murphy2019relational,zang2020moflow,liu2019graph}. GraphITE theoretically requires this property although it does not hold for most practical GNNs; however, a non-invertible GNN shows satisfactory performance in practice,  as shown in the experimental section.

\vspace{3mm}
Recently, several studies have considered causal inference in \emph{graph-structured input domains}~\cite{guo2020learning,alvari2019less,veitch2019using,Harada2020CF,ma2020causal}, where the input space has a graph structure representing proximal relations among target individuals.
However, to the best of our knowledge, no study has explored treatment effect estimation with \emph{graph-structured treatments}, which is at the  intersection of the above two topics of practical importance.

\section{Problem Definition}
We consider the problem of estimating the outcomes of treatments with graph structures from biased observational data.
Let $\mathcal{D}=\{(x_i, t_i, y_{i}^{t_i})\}^{N}_{i=1}\in \mathcal{X}\times\mathcal{T}\times\mathcal{Y}$  be a biased observational dataset, where $x_i\in\mathcal{X}$ is the covariate vector of the $i$-th target individual, 
$t_i \in \mathcal{T}$ is the treatment performed on the target individual, and $y_{i}^{t_i}$ is the outcome.\footnote{Owing to the counterfactual nature of the problem, we are unable to observe the outcomes of the other treatments as they are not performed.}
We assume the covariate space $\mathcal{X}=\mathbb{R}^D$, treatment space $\mathcal{T}=\{1,2,\ldots,|\mathcal{T}|\}$, and outcome space $\mathcal{Y}=\mathbb{R}$.
In addition to $\mathcal{D}$, we assume each treatment $j \in \mathcal{T}$ is associated with a graph $G_j =(\mathcal{V}_j, \mathcal{E}_j)$, where $\mathcal{V}_j$ denotes a set of nodes and $\mathcal{E}_j \subseteq \mathcal{V}_j \times \mathcal{V}_j$ denotes a set of edges. We denote the set of the treatment graphs by $\mathcal{G}=\{G_j\}_{j=1}^{|\mathcal{T}|}$.
Our goal is to, given $\mathcal{D}$ and $\mathcal{G}$, estimate an outcome prediction function $f: \mathcal{X} \times \mathcal{T}\rightarrow \mathcal{Y}$.

Figure~\ref{fig:intro} illustrates our problem setting in the context of medical treatments.
In the observational data, there are multiple drugs that could be applied to the target individual, where each drug corresponds to a treatment and is associated with a graph representing its molecular structure. Only the outcome $y_i^{t_i}$for the actual prescribed drug $t_i$ (i.e., the factual outcome) is observed, and those for the other drugs (i.e., counterfactual outcomes) are not observed.
Because a doctor prescribes a drug based on the condition of the target patient $x_i$, there is a bias in the choice of $t_i$ in the observational data.

A potential difficulty in our problem is that the number of treatments can be large, say $|\mathcal{T}|>100$; it is clear that this can cause a data scarcity issue. 
In our scenario, graphs are available as auxiliary information for the treatments, which potentially help in dealing with such a large number of treatments.

Following the existing work, we make the typical assumptions in the Rubin-Neyman framework~\cite{rubin2005causal}:
(i) Stable unit treatment value; the outcome of each instance is not affected by the treatments assigned to other instances. (ii) Unconfoundedness; the treatment assignment to an instance is independent of the outcome given the covariates (i.e., the confounder variables). (iii) Overlap; each instance has a positive probability of treatment assignment, i.e., $\forall  x, t$, $p(x,  t)>0$.

\section{GraphITE}
Previous studies on individual treatment effect estimation have not considered rich information associated with treatments, which in our case is given as graphs.
We expect that the use of such auxiliary information will be effective, especially when the number of treatments is relatively high and the training dataset is biased.
We propose GraphITE, which utilizes graph-structured treatments while reducing selection bias effectively.
We first introduce the network architecture of GraphITE, and then apply HSIC regularization to estimate outcomes appropriately from a biased dataset.

\subsection{Model}

\begin{figure}[tb]
 \centering
   \begin{center}
    \includegraphics[width=1\linewidth, height=28mm]{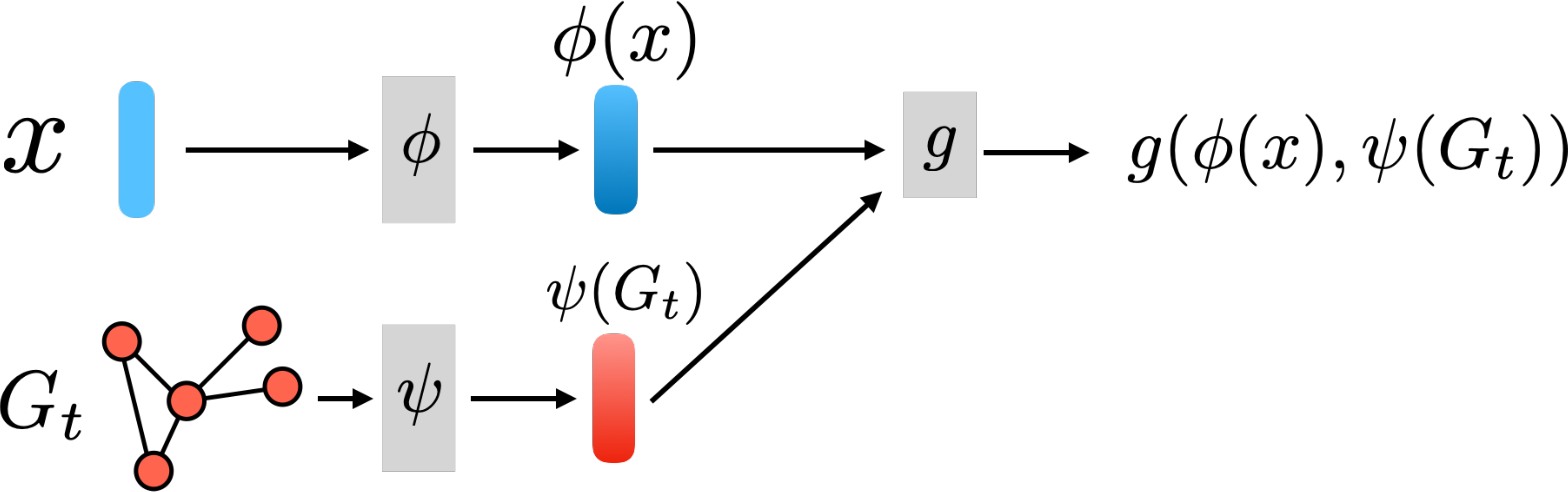}
       \centering
 \end{center}
  \caption{\small{The model architecture of GraphITE. A target individual $x$ and graph-structured treatment $G_t$ are the inputs. The $\phi$ and $\psi$ map them to the low-dimensional representations, where $\phi$ is a standard feed-forward neural network and $\psi$ is a graph neural network. The two representation vectors $\phi(x)$ and $\psi(G_t)$ are concatenated to be an input to another feed-forward network $g$, which predicts the outcome.
 }}\label{fig:model_vanilla}
\end{figure}

The model of GraphITE consists of three components: two mapping functions $\phi:\mathcal{X}\rightarrow \Phi$ and $\psi:\mathcal{T}\rightarrow \Psi$ for extracting representations of the input and treatment graph, respectively, and a prediction function $g:\Phi \times \Psi\rightarrow\mathcal{Y}$ for predicting the outcome,
where $\Phi$ and $\Psi$ are the latent representation spaces of the inputs and treatments induced by $\phi$ and $\psi$, respectively.
Figure~\ref{fig:model_vanilla} illustrates the overview of the neural network architecture of GraphITE.

As mentioned earlier, the mapping function $\psi$ extracts representations that capture the features of graph-structured treatments.
If we simply take $\psi$ as a one-hot encoding of discrete treatments, it coincides with the standard setting with multiple treatments; however, this approach cannot take advantage of the rich structural information that the graph treatments have, and therefore suffers from a large number of treatments.

As the mapping function $\psi$ of the treatment graphs, we employ a GNN. 
GNNs have been successfully applied in various domains~\cite{duvenaud2015convolutional,kipf2016semi,gilmer2017neural} and 
are capable of extracting features of graphs owing to the flexible expressive power of neural networks optimized in an end-to-end manner.

The representation vector of graph-structured treatment $G=(\mathcal{V},\mathcal{E})$ is otained as follows.
First, for each node $v_k \in \mathcal{V}$, the representation of $v_k$ is initialized to a low-dimensional vector $\mathbf v_k^{(0)} \in \mathbb R^{D_\Psi}$ determined by randomized initialization depending on the node label, such as the atom type.
At the $c$-th layer of the GNN, the node representations are updated using
\begin{eqnarray}
\label{eq:graph_convolution}
\mathbf v_k^{(c)} = \sigma^{\mathcal{V}} \left( \mathbf W \mathbf v_k^{(c-1)}
+ \sum_{v_m \in \mathcal{N}_k} \mathbf M \mathbf v_m^{(c-1)} \right),
\end{eqnarray}
where $\sigma^{\mathcal{V}}$ is an activation function, such as the ReLU function, $\mathcal{N}_k$ is the set of nodes adjacent to $v_k$,
and $\mathbf W$ and $\mathbf M$ are transformation matrices.
After the updates through $C$ layers, the representations of all the nodes are aggregated into a graph-level representation $\psi(G)$ as
\begin{equation}
\label{eq:representation}
\psi(G)= \sum_{v_k \in \mathcal{V}} \sigma_G \left( \sum_{c=0}^{C} \mathbf v_k^{(c)}\right),
\end{equation}
where $\sigma_G$ is an activation function, such as the softmax function.

Note that, as we will see later, we require the treatment mapping function to be invertible, i.e., one-to-one, which most GNNs are not; however, some recent studies have proposed GNNs with the one-to-one property~\cite{murphy2019relational,zang2020moflow,liu2019graph}. In our experiments, we use a non-invertible GNN  which exhibits satisfactory performance.

  \begin{figure}[tb]
 \centering
   \begin{center}
    \includegraphics[width=\linewidth, height=36mm]{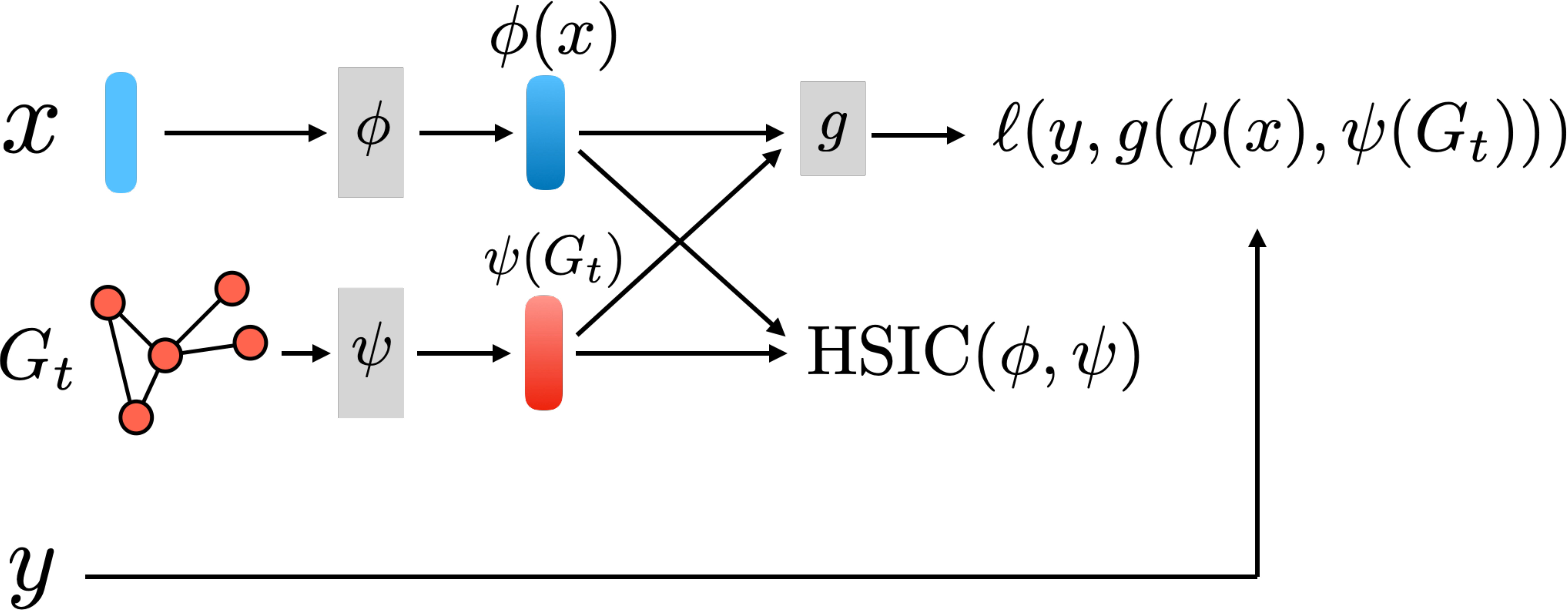}
       \centering
\end{center}
  \caption{\small{Training of GraphITE using the HSIC regularization. 
  In addition to the prediction loss function $\ell$ between the prediction $g(\phi(x),\psi(G_t))$ and the true outcome $y$, the HSIC regularization term encourages the two representations $\phi(x)$ and $\psi(G_t)$ to be independent of each other in order to mitigate selection biases. 
 Theorem~\ref{theorem:hsicbound} gives a theoretical guarantee on how the HSIC regularization contributes to the bias mitigation.} }\label{fig:model_hsic}
 \end{figure}

\subsection{Bias mitigation using HSIC regularization}
With an unbiased dataset collected through randomized controlled trials (RCT), it suffices to minimize the objective function
\begin{equation}\label{eq:supervised}
\sum_{i=1}^{N} \ell(y_i^{t_i}, g(\phi(x_i), \psi(G_{t_i}))
\end{equation}
to estimate the components of GraphITE, $\phi, \psi$, and $g$,
where $\ell$ is a loss function, such as mean squared error~(MSE).
However, the objective function is biased when the training dataset is biased, and we must adjust it to mitigate the negative effect. 

We first propose our approach from an intuitive viewpoint.
The main source of the bias is that, in contrast with RCT, the treatments in the observational data are selected depending on the target individuals (i.e., the covariates).
Our idea for mitigating the bias is to reduce the dependency, i.e., to find representations of the target and treatment that are as independent of each other as possible.
To implement this idea, we employ HSIC~\cite{gretton2008kernel} to measure the independence; the HSIC is defined as 
\begin{equation}\label{eq:hsic}
\text{HSIC}(\phi, \psi) =(N-1)^{-2}\text{tr}
({\bf K}^{\Phi}{\bf H}{\bf K}^{\Psi}{\bf H}),
\end{equation}
where ${\bf K}^{\Phi}$ and ${\bf K}^{\Psi}$ are the kernel matrices of the representations of the targets and treatments, respectively, and ${\bf H}$ is the centering matrix ${\bf H}=\mathbf{I}-\frac{1}{N}\mathbf{1}$. 
If the kernel function is characteristic, HSIC becomes $0$ in expectation if and only if the two representations are independent; we use the Gaussian kernel as the kernel function. 
HSIC is somewhat computationally expensive, which requires $\mathcal{O}(N^{2})$ time and space complexity, and does not scale to the sample size. 
Therefore, for the sake of computational convenience, we compute the HSIC loss in a mini-batch manner.

With the HSIC as a regularization term, our objective function is modified to 
\begin{equation}\label{eq:loss}
\sum_{i=1}^{N} \ell(y_i^{t_i}, g(\phi(x_i), \psi(G_{t_i})) + \lambda~\text{HSIC}(\phi, \psi),
  \end{equation}
where $\lambda$ is the regularization hyper-parameter. 
The $\phi$ is implemented as a standard feed-forward neural network, while $\psi$ is a GNN; the $\phi$ and $\psi$ are concatenated as an input to $g$ that is another feed-forward network. Figure~\ref{fig:model_hsic} illustrates the training of GraphITE using the HSIC regularization.

The objective function (\ref{eq:loss}) is optimized in a mini-batch manner using Adam~\cite{kingma2014adam};
more specifically, each epoch divides the entire training dataset $\mathcal{D}$ into mini batches without overlapping, and approximates the loss function and HSIC term with them.
Recent theoretical analysis reveals that minimizing the HSIC loss in a mini-batch manner is equivalent to bagging block HSIC method~\cite{nguyen2021wide,yamada2018post}, which ensures that it converges to the same value. 
The training procedure for GraphITE is outlined in Algorithm~\ref{alg:train}.

Finally, we give some historical remarks explaining why
we specifically chose the HSIC as the regularization term.
Previous studies have considered a broad class of regularization terms, integral probability metrics (IPM)~\cite{johansson2016learning,shalit2017estimating}; 
however, they basically assume a binary treatment or a relatively small number of treatments, and they cannot be directly applied to a large number of treatments. For example, typical IPM such as the maximum mean discrepancy~(MMD) and Wasserstein distance, require many regularization terms for all pairs of treatments; otherwise, an expedient ``pivot" control treatment must be set, which is not effective, as demonstrated in our experiments.
The use of the HSIC is proposed by Lopez et al.~\cite{lopez2020cost} as it naturally allows multiple treatments. 
However, they did not consider learning representations of treatments.

  \begin{algorithm}[!tb]
  \caption{GraphITE training procedure}\label{alg:train}
  \SetAlgoLined
 {\bf Input:} Observational data:~$\mathcal{D}=\{(x_i, t_i, y^{t_i}_i)\}^{N}_{i=1}\sim p^\text{train}$, \\
 a set of graph-structured treatments $\mathcal{G}=\{G_j\}_{j=1}^{|\mathcal{T}|}$, \\
 and a hyperparameter $\lambda \geq 0$.\\
 {\bf Output:} An outcome prediction model $f=(g,\phi,\psi)$. \\
  \While{\text{not converged}}{
   Sample a mini batch  $\mathcal{B}=\{(x_{i_o}, t_{i_o}, y^{t_{i_o}}_{i_o})\}^{B}_{o=1} \subset \mathcal{D}$\\
  {\# Mini-batch approximation of the supervised loss~(Eq.~(\ref{eq:supervised}))}\\
Compute $L_\mathcal{B}(g,\phi,\psi) = \sum_{o=1}^{B}\ell(y_{i_o}, g(\phi(x_{i_o}),\psi(G_{t_{i_o}})))$\\
 {\#  Mini-batch approximation of the HSIC loss~(Eq.~(\ref{eq:hsic}))}\\
Compute $\text{HSIC}_\mathcal{B}(\phi,\psi) = \text{HSIC}(\phi_{x\in\mathcal{B}}, \psi_{t\in\mathcal{B}})$\\
  {\#  Update the parameters of $f$}\\
Minimize 
$L_\mathcal{B}(g,\phi,\psi) +\lambda\cdot \text{HSIC}_\mathcal{B}(\phi,\psi)$
using SGD
 }
\end{algorithm}

\subsection{Theoretical justification of HSIC regularization}

Now we consider the theoretical justification for using the HSIC regularization in GraphITE.
Our discussion is based on generalizations of the theories~\cite{lopez2020cost,johansson2020generalization} to treatments with features. 
We also discuss the benefits of our formulation and how it makes the prediction model more flexibly deal with complex situations than the existing approaches.

Denote by $p^\text{train}$ the probability distribution on $\mathcal{X}\times\mathcal{T}$ the training dataset $\mathcal{D}$ follows, and by $p^\text{test}$ the one  for the test dataset.
We assume that the test distribution has the form of $p^{\text{test}}(x,t)=p^{\mathcal{X}}(x)p^{\mathcal{T}}(t)$ because we want our model to perform well on the distribution where treatments do not depend on the covariates.

For some (unknown target) function $f^*: \mathcal{X} \times \mathcal{T}\rightarrow\mathcal{Y}$ and probability distribution $p$ over $\mathcal{X} \times \mathcal{T}$, let the expected risk of our prediction model $f: \mathcal{X} \times \mathcal{T}\rightarrow\mathcal{Y}$ with the mapping functions $\phi$, $\psi$ and a predictive function $g$ be 
\begin{eqnarray}
\epsilon_{p}(f_{g,\phi,\psi})=\mathbb{E}_{(x,t)\sim {p}} [\ell(f( g(\phi(x),\psi(G_t))), f^*(x, t))  ].
\end{eqnarray}

Then we have the following theoretical upper bound of the expected risk on the test distribution in terms of that for the training distribution and the HSIC regularization term.
\begin{theorem}\label{theorem:hsicbound}{
Let $\epsilon_{p^\text{train}}(f_g)$ and $\epsilon_{p^\text{test}}(f_g)$ be the expected risk for the training distribution and the test distribution, respectively.
Let the IPM~between two distributions $p$ and q with respect to a function family $\mathcal{H}$ be
\begin{align}
\text{IPM}_{\mathcal{H}}(p,q)=\sup_{h\in\mathcal{H}}|\mathbb{E}_{p}[h]-\mathbb{E}_{q}[h]|.
\end{align}
Let $J_\phi^{-1}(\xi)$, $J_{\psi}^{-1}(\tau)$ be the Jacobian matrices of $\phi^{-1}$ and $\psi^{-1}$ at $\xi$ and $\tau$, respectively.  
Assume that there exist positive constants $A$ and $B$ that satisfy  $|J_{\phi^{-1}(\xi)}||J_{\psi^{-1}(\tau)}|\leq A$  and $\frac{\ell_{f_{g,\phi,\psi}}}{B}\in\mathcal{H}\subseteq\{g:\Phi\times\Psi\rightarrow\mathcal{Y}\}$.
Then the expected risk for the test distribution is upper-bounded by %
\begin{align}\label{HSICbound}
\epsilon_{p^\text{test}}(f_{g,\phi,\psi})
\leq \epsilon_{p^\text{train}}(f_{g,\phi,\psi}) + \lambda \cdot \text{HSIC}(\phi, \psi).
\end{align}}
\begin{proof}
\begin{align*}
&\hspace{-1mm} \epsilon_{p_\text{test}}(f_{\phi, \psi})-\epsilon_{p_\text{train}}(f_{g,\phi,\psi})\\=&\int_{\mathcal{X}\times\mathcal{T}} \hspace{-1mm} \ell_{f_{g,\phi,\psi}}(x,t)(p(x)p(t))dxdt-\int_{\mathcal{X}\times\mathcal{T}}\hspace{-1mm} \ell_{f_{g,\phi,\psi}}(x,t)p(x,t)) dxdt\\
=&\int_{\mathcal{X}\times\mathcal{T}}\ell_{f_{g,\phi,\psi}}(x,t)(p(x)p(t)-p(x,t)) dxdt\\
=&\int_{\Phi\times\Psi}\ell_{f_{g,\phi,\psi}}(\phi^{-1}(\xi), \psi^{-1}(\tau))(p(\xi)p(\tau) -p(\xi,\tau))\notag\\&\hspace{8mm}\cdot|J_{\phi^{-1}(\xi)}||J_{\psi^{-1}(\tau)}|d\xi d\tau\\
\leq&A\int_{\Phi\times\Psi} \ell_{f_{g,\phi,\psi}}(\phi^{-1}(\xi), \psi^{-1}(\tau))(p(\xi) p(\tau) -p(\xi,\tau))d\xi d\tau \\
\leq&A\cdot B\sup_{g\in\mathcal{H}} \left|\int_{\Phi\times\Psi} \hspace{-1mm} g(\phi^{-1}(\xi), \psi^{-1}(\tau))(p(\xi)p(\tau) -p(\xi,\tau))d\xi d\tau\right| \\
\leq&A\cdot B\cdot \text{IPM}_{\mathcal{H}}(p(\xi)p(\tau), p(\xi,\tau)) \\
\leq&\cdot A\cdot B\cdot C\cdot \text{HSIC}(p(\xi), p(\tau)),
\end{align*}
where $\ell_{f_{g,\phi,\psi}}=\mathbb{E}_{y}[\ell(y^{t}, g(\phi(x), \psi(G_{t}))\mid x, t ]$.
$C$ is a constant that denotes the radius of the function space.
By setting $\lambda = A \cdot B\cdot C$, we obtain the inequality (\ref{HSICbound}).
\end{proof}
\end{theorem}
The theorem states that minimizing the HSIC between the representations of the targets and graph-structured treatments leads to minimizing the expected risk for the test distribution, making the predictive model to handle even unseen graph-structured treatments unbiasedly. 
For  the inequality (\ref{HSICbound}) to hold, we require several conditions: $\phi$ and $\psi$ must be twice-differentiable one-to-one mapping functions, and the HSIC must be defined using continuous, bounded, positive semi-definite kernels $k^\Phi:\Phi\times\Phi\rightarrow\mathbb{R}$ and $k^\Psi:\Psi\times\Psi\rightarrow\mathbb{R}$.
The $\lambda$ must  also be theoretically determined based on the radius of the function space in which $f$ lies, but empirically, we simply treat it as a hyper-parameter. Our choices of the kernels and the hyper-parameters are detailed in Section~\ref{sec:experiments}. 

Note that MMD and Wasserstein distance are special cases of IPM when the function family includes the set of $1$-Lipschitz functions and the set of unit norm functions in a universal reproducing norm Hirbert space~\cite{shalit2017estimating}, respectively. Hence, they are obviously bounded by the inequality~(\ref{HSICbound}).

  \begin{table}[tb]
\centering
 \caption{\small{Summary statistics of the datasets.}}\label{table:data}
\begin{tabular}{|c|c|c|c|}
\hline
{\bf Dataset} & {\bf \#Units} &{\bf \#Treatments}  & {\bf \#Interactions} \\ \hline
CCLE    &  $491$           &         $24$     &    $11{,}054$          \\ \hline
GDSC    &   $925$          &      $117$        &   $105{,}694$          \\ \hline
\end{tabular}
\end{table}

\section{Experiments}\label{sec:experiments}

We experimentally investigate the performance of the proposed GraphITE and its merits of using the GNN and the HSIC regularization compared with various baseline methods on two real-world datasets.

\subsection{Datasets}
We use two real-world datasets on drug responses: the Cancer Cell-Line Encyclopedia (CCLE) dataset~\cite{barretina2012cancer} and  Genomics of Drug Sensitivity in Cancer (GDSC) dataset~\cite{yang2012genomics}. 
Table~\ref{table:data} lists their basic statistics. \#Units, \#Treatments, and \#Interactions represent the number of units, the number of treatments, and the number of labeled data in a dataset.
CCLE is a relatively small dataset with a moderate number of treatments, while GDSC has more than 100 treatments.
Both of the datasets include $\text{IC}_{50}$ values for drug--cell pairs, which are known to be closely related to drug sensitivity.
Namely, we define the drug sensitivity as $y=-\log\text{IC}_{50}$  following previous studies~\cite{lind2019predicting,suphavilai2018predicting}, which is the regression target in our experiments.
We use the similarity matrices of each cell line as the input features.
Both datasets are publicly available\footnote{\url{https://github.com/CSB5/CaDRReS}}~\cite{suphavilai2018predicting}.

Because the two original datasets are fairly close to complete observations (specifically, their observation rates are about $94\%$ and $98\%$, respectively), 
we simply assume that they are complete, and introduce synthetic observation biases to extract biased training datasets, and then test the predictors obtained from them on the remainder.
We introduce synthetic treatment bias that assigns treatment $t$ using $t\sim \text{Categorical}(\text{softmax}(\rho y))$ following previous studies~\cite{schwab2019learning,schwab2018perfect,bica2021real}.
The $\rho=\frac{\eta}{100\sigma}$ is a bias coefficient, where $\eta$ is the magnitude of selection bias and $\sigma$ is the standard deviation of target values.  A larger $\eta$ indicates a higher selection probability; intuitively, this indicates that scientists are more likely to conduct experiments for drug--cell pairs with higher sensitivity values. 
Note that although this bias generation procedure does not necessarily satisfy the typical unconfoundedness assumption, it is more practical and reasonable in the sense that the scientists likely to select promising experimental targets based on their knowledge and experience. In other words, we assume scientists are not incompetent.

  \begin{table*}[tb]
    \caption{\small{Performance comparison of different methods on the CCLE and  GDSC dataset in terms of RMSE and CI. $^\dagger$ and $\ddagger$ indicate statistically significantly better performance of the proposed GraphITE than the baseline by the paired $t$-test with $p<0.05$ and $p<0.01$, respectively. The bold results indicate the statistically significant best results. The shaded rows indicate the GNN-based methods. Lower RMSEs are better, and higher CIs are  better.}}\label{table:results}

\centering
\begin{tabular}{ccccccccc}
\toprule[2pt]
                & \multicolumn{2}{c}{ \bf{CCLE} }        &              &  \multicolumn{2}{c}{ \bf{GDSC} } \\[3pt] \cmidrule{1-6}

Method                       & RMSE & CI & \multicolumn{1}{l}{} &  RMSE &  CI  \\[3pt] \cmidrule{1-6}
\multicolumn{1}{c}{Mean }    &        $^\ddagger3.777_{\pm{0.101}}$       &       $-$         && $^\ddagger 4.030_{\pm 0.102}$   & $-$  \\[2pt]

\multicolumn{1}{c}{OLS }    &      $^\ddagger 4.861_{\pm{0.755}}$          &      $^\ddagger 0.642_{\pm{0.021}}$          && $^\ddagger6.463_{\pm{0.493}}$   &  $^\ddagger 0.602_{\pm{0.018}}$  \\[2pt]
\multicolumn{1}{c}{BART }    &      $^\ddagger 2.993_{\pm{0.203}}$          &      $^\ddagger 0.711_{\pm{0.016}}$          && $^\ddagger3.965_{\pm{0.102}}$   &  $^\ddagger 0.632_{\pm{0.015}}$  \\[2pt]
\cmidrule{1-6}
 Treatment Embedding &        $^\ddagger2.662_{\pm{0.161}}$       &       $^\ddagger0.724_{\pm{0.013}}$        &  &       $^\ddagger3.642_{\pm{0.131}}$        &       $^\ddagger0.670_{\pm{0.015}}$        \\[2pt]
\multicolumn{1}{c}{TARNet }    &        $^\ddagger2.831_{\pm{0.123}}$       &       $^\ddagger 0.711_{\pm{0.013}}$         && $^\ddagger3.813_{\pm 0.135}$   & $^\ddagger  0.663_{\pm 0.009}$  \\[2pt]
 \multicolumn{1}{c}{CFR }    &        $^\ddagger2.822_{\pm{0.121}}$       &       $^\ddagger 0.712_{\pm{0.013}}$         && $^\ddagger 3.792_{\pm 0.134}$   & $^\ddagger  0.664_{\pm 0.009}$  \\[2pt]
  \multicolumn{1}{c}{GANITE }    &        $^\ddagger3.652_{\pm{0.211}}$       &       $^\ddagger 0.651_{\pm{0.023}}$         && $^\ddagger 7.739_{\pm 1.394}$   & $^\ddagger  0.613_{\pm 0.018}$  \\[2pt]
\cmidrule{1-6}
\rowcolor{Gray}
\rowcolor{Gray}
 GNN  &        $^\ddagger2.652_{\pm{0.123}}$       &       $^\ddagger0.720_{\pm{0.010}}$        &  &       $^\ddagger3.553_{\pm{0.126}}$        &       $^\ddagger0.681_{\pm{0.010}}$        \\[2pt]
\rowcolor{Gray}
 GNN+MMD  &        $^\dagger2.596_{\pm{0.162}}$       &       $^\dagger0.726_{\pm{0.014}}$        &  &       $^\ddagger3.531_{\pm{0.136}}$        &       $^\ddagger0.683_{\pm{0.013}}$        \\[2pt]
\rowcolor{Gray}
 GraphITE~(Proposed)  &        $\bf 2.561_{\pm{0.112}}$       &       $\bf 0.732_{\pm{0.009}}$        &  &       $\bf 3.421_{\pm{0.135}}$        &       $\bf 0.695_{\pm{0.015}}$        \\
\bottomrule[2pt]
    \end{tabular}
\end{table*}

 \begin{figure*}[tb]
\centering
\begin{minipage}{0.245\hsize}
   \includegraphics[width=\linewidth]{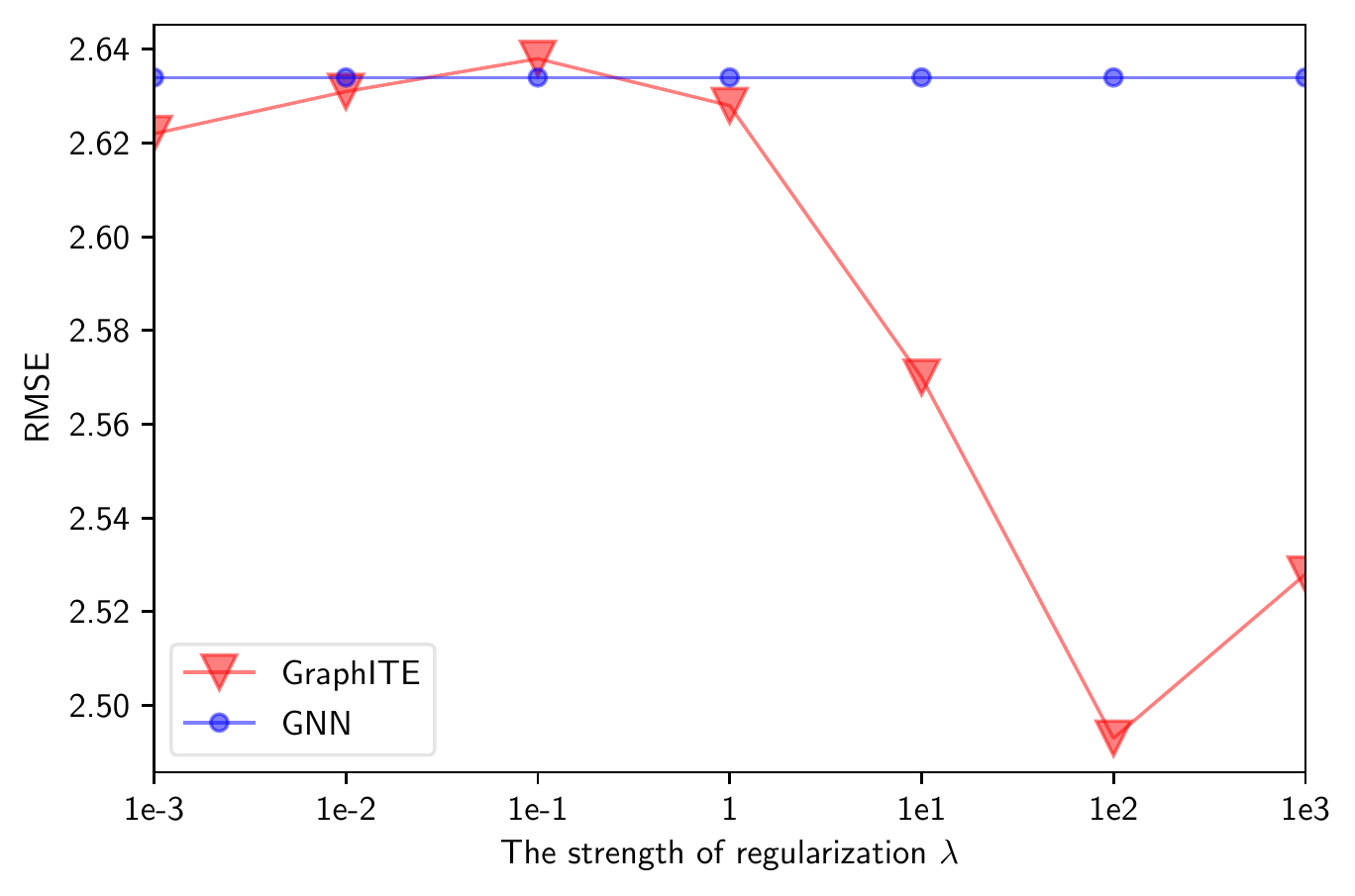}
      \centering
   (a): RMSE~(CCLE)
 \end{minipage}
 \begin{minipage}{0.245\hsize}
   \includegraphics[width=\linewidth]{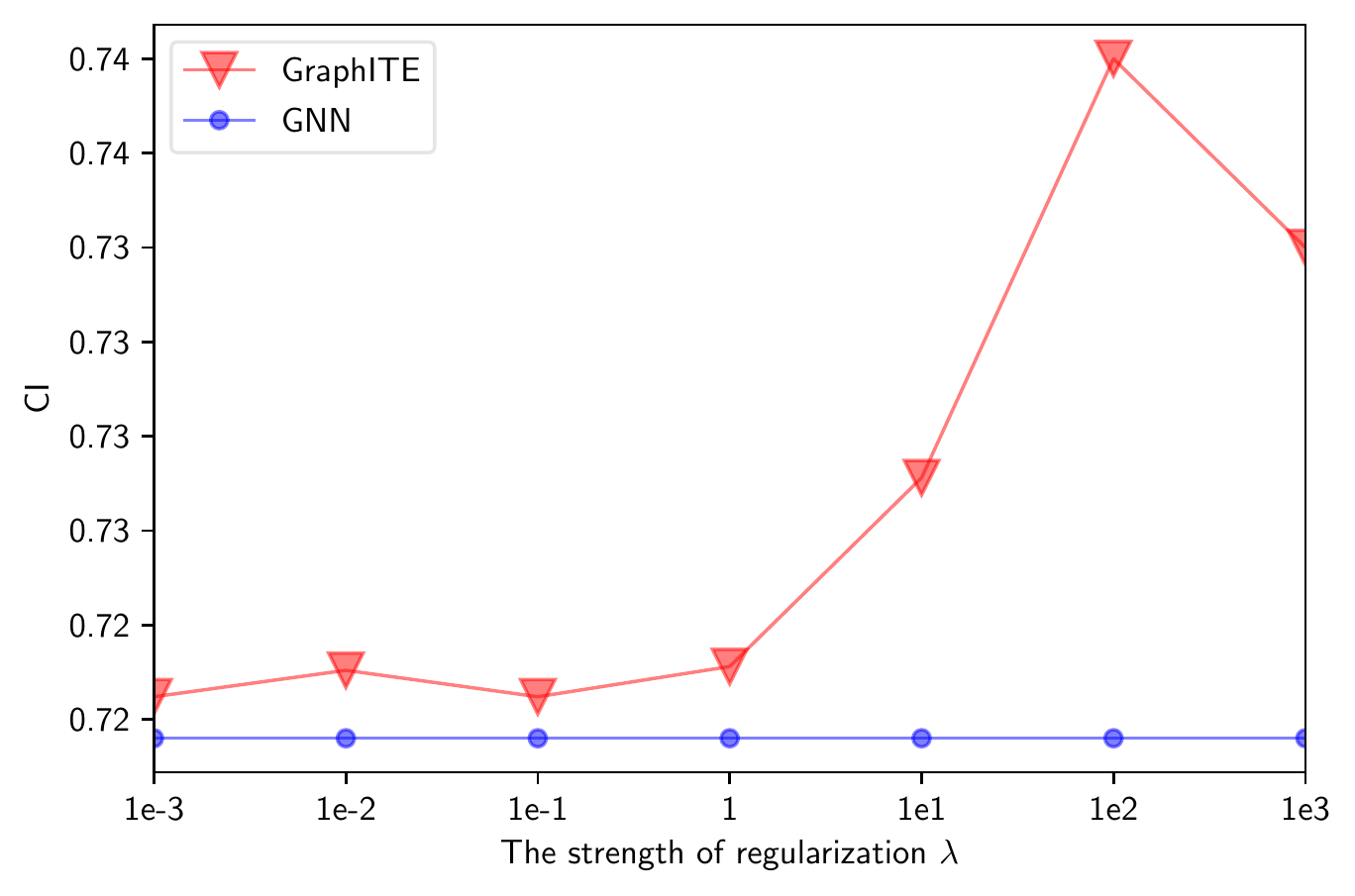}
      \centering
   (b): CI~(CCLE)
 \end{minipage}
 \begin{minipage}{0.245\hsize}
   \includegraphics[width=\linewidth]{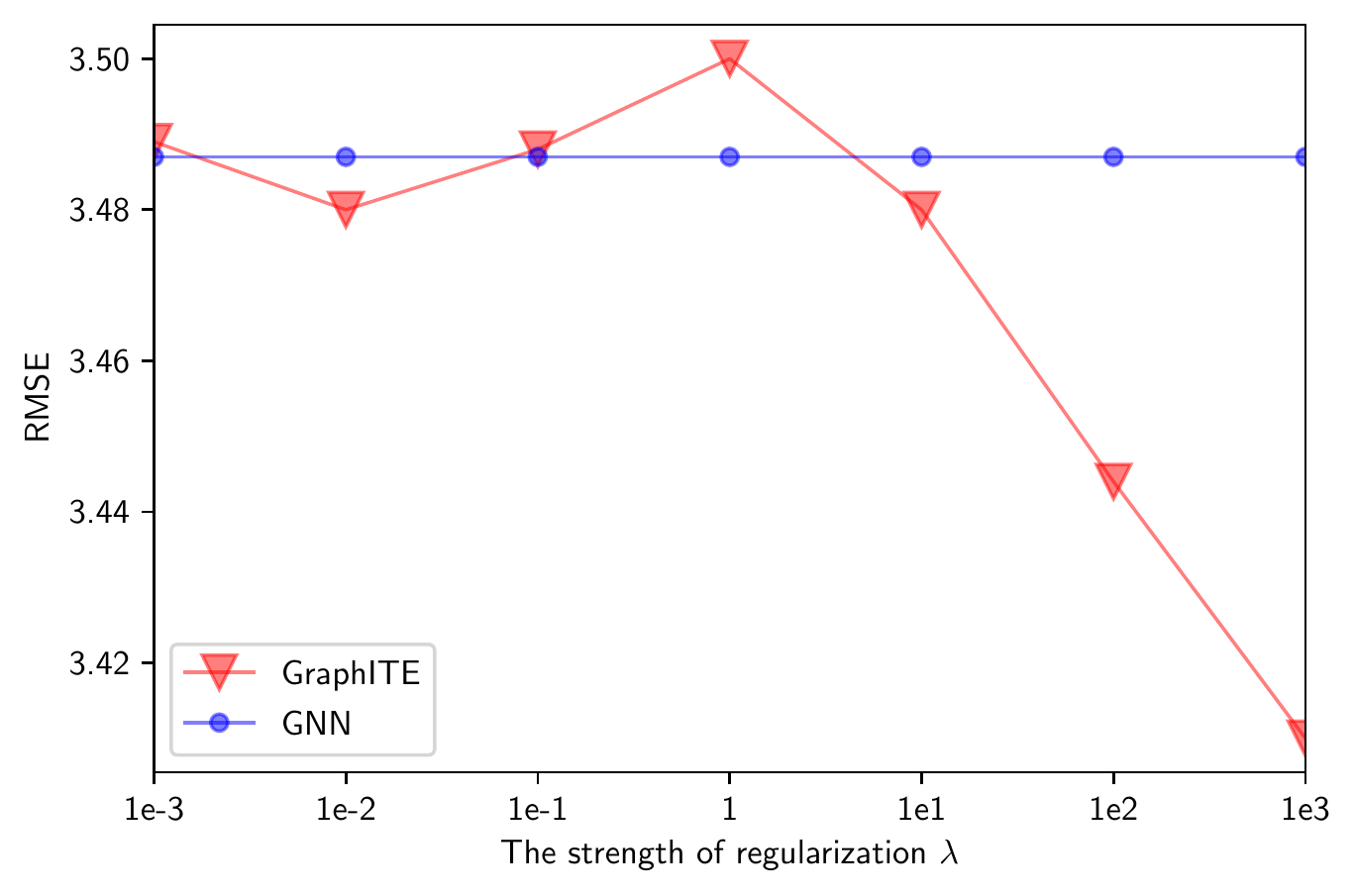}
      \centering
   (c): RMSE~(GDSC)
 \end{minipage}
 \centering
  \begin{minipage}{0.245\hsize}
   \includegraphics[width=\linewidth]{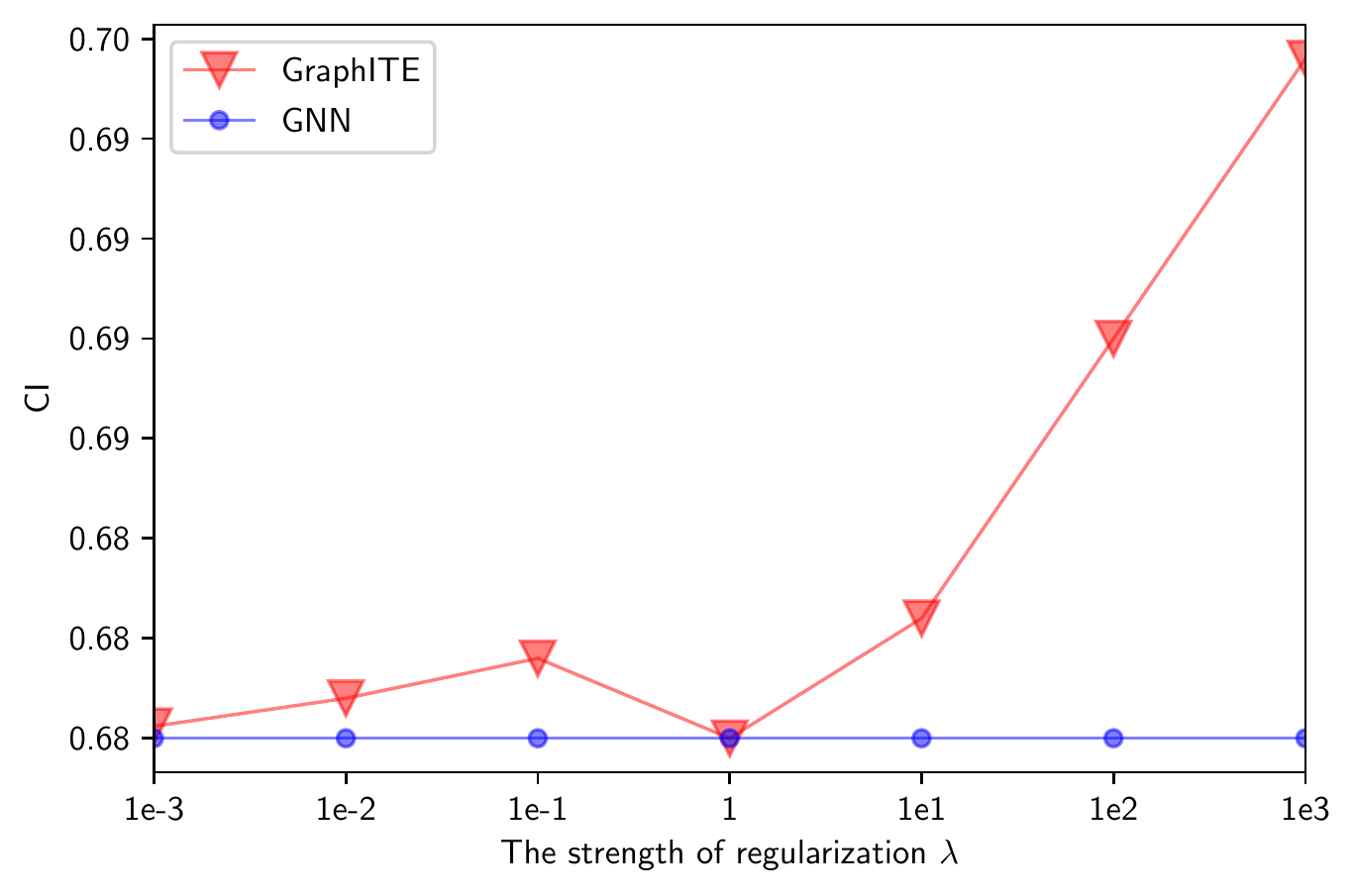}
      \centering
   (d): CI~(GDSC)
 \end{minipage}\\
 \caption{\small{Sensitivity of the results to the regularization strength $\lambda$. Although the optimal choices significantly improve the performance, at worst the other choices do not harm the performance. 
 }}\label{fig:reg}
\end{figure*}

 \begin{figure*}[tb]
\centering
\begin{minipage}{0.245\hsize} %
   \includegraphics[width=\linewidth]{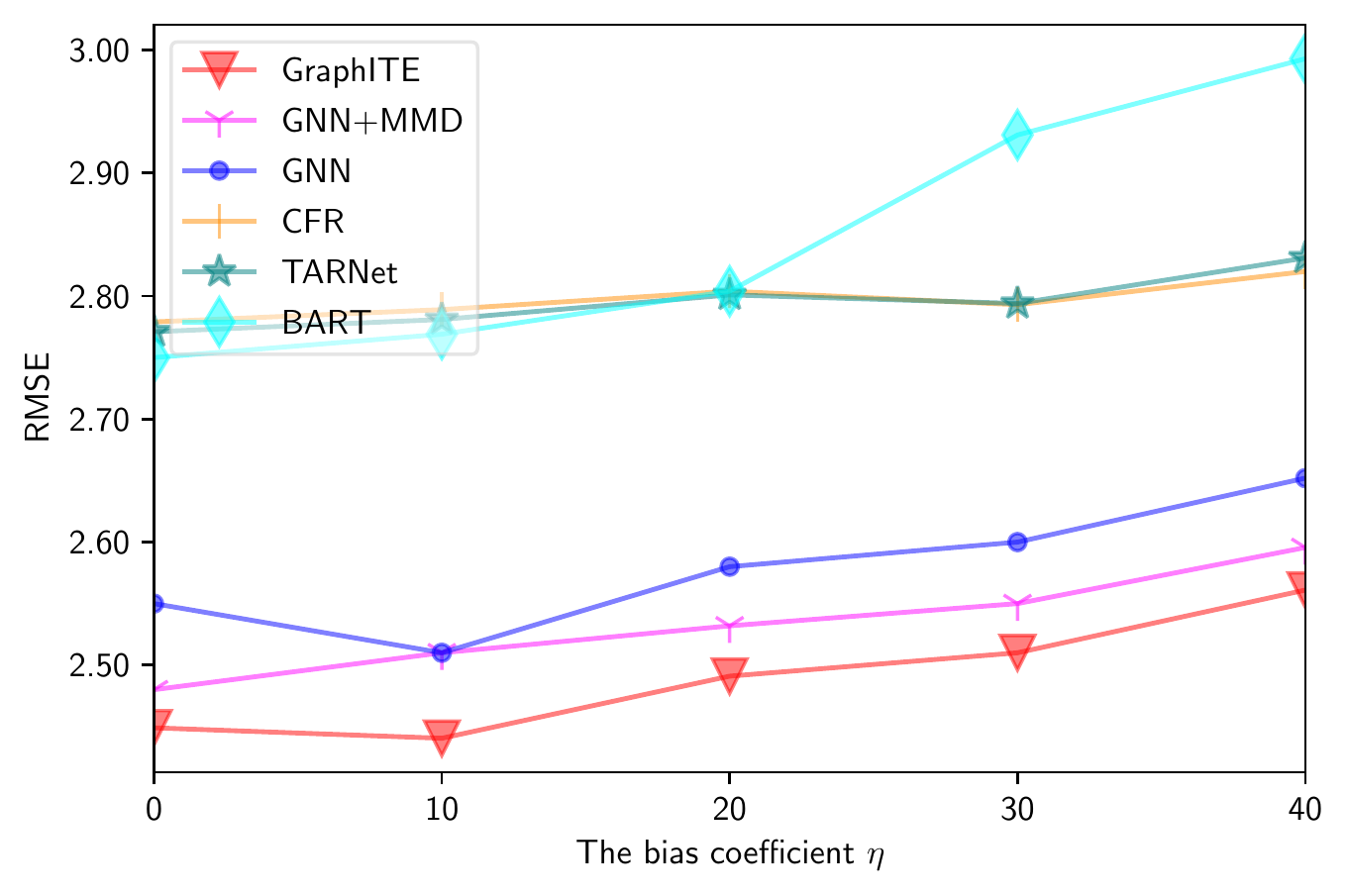}
      \centering
   (a): RMSE~(CCLE)
 \end{minipage}
 \begin{minipage}{0.245\hsize}
   \includegraphics[width=\linewidth]{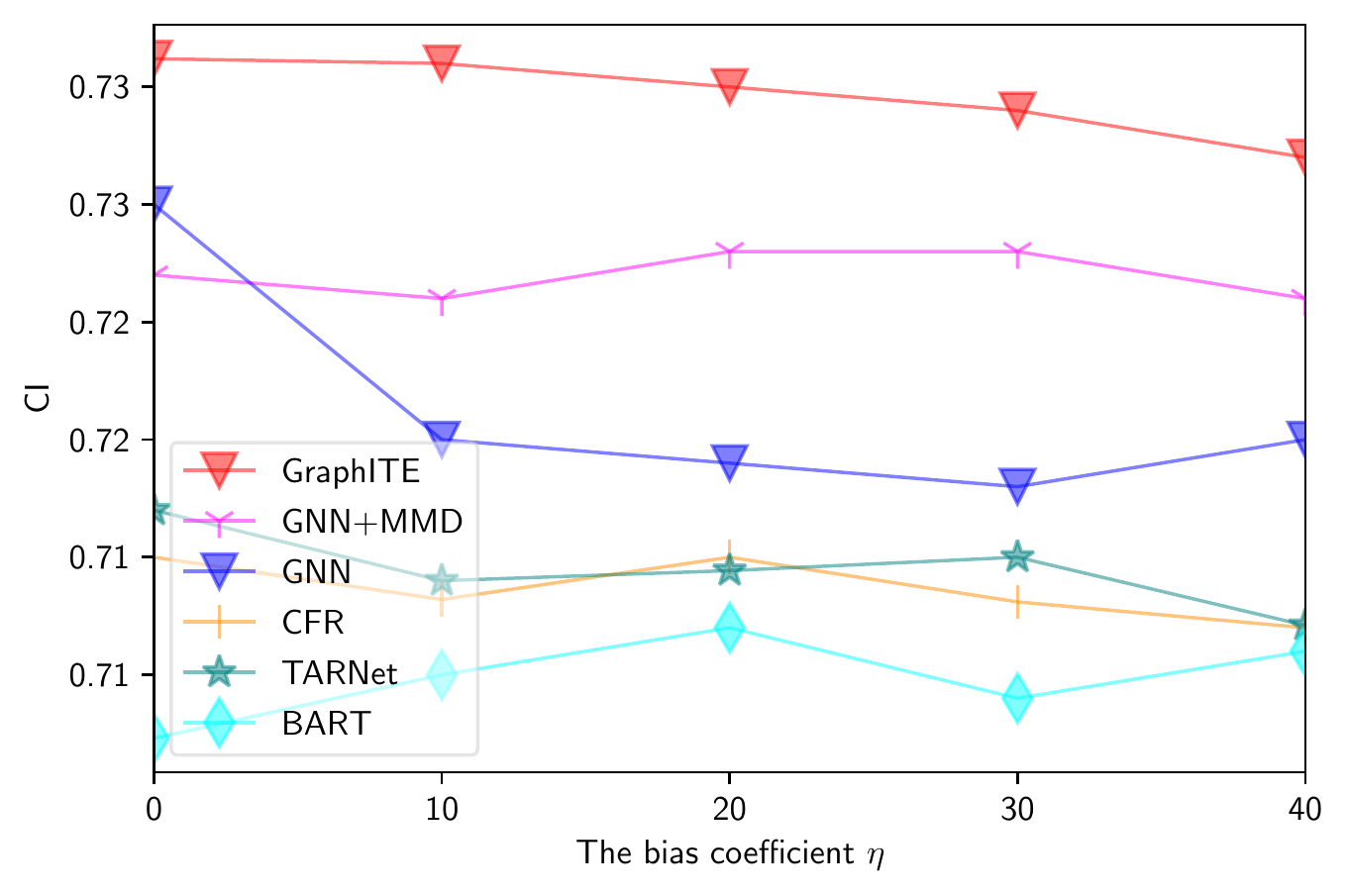}
      \centering
   (b): CI~(CCLE)
 \end{minipage}
 \begin{minipage}{0.245\hsize}
   \includegraphics[width=\linewidth]{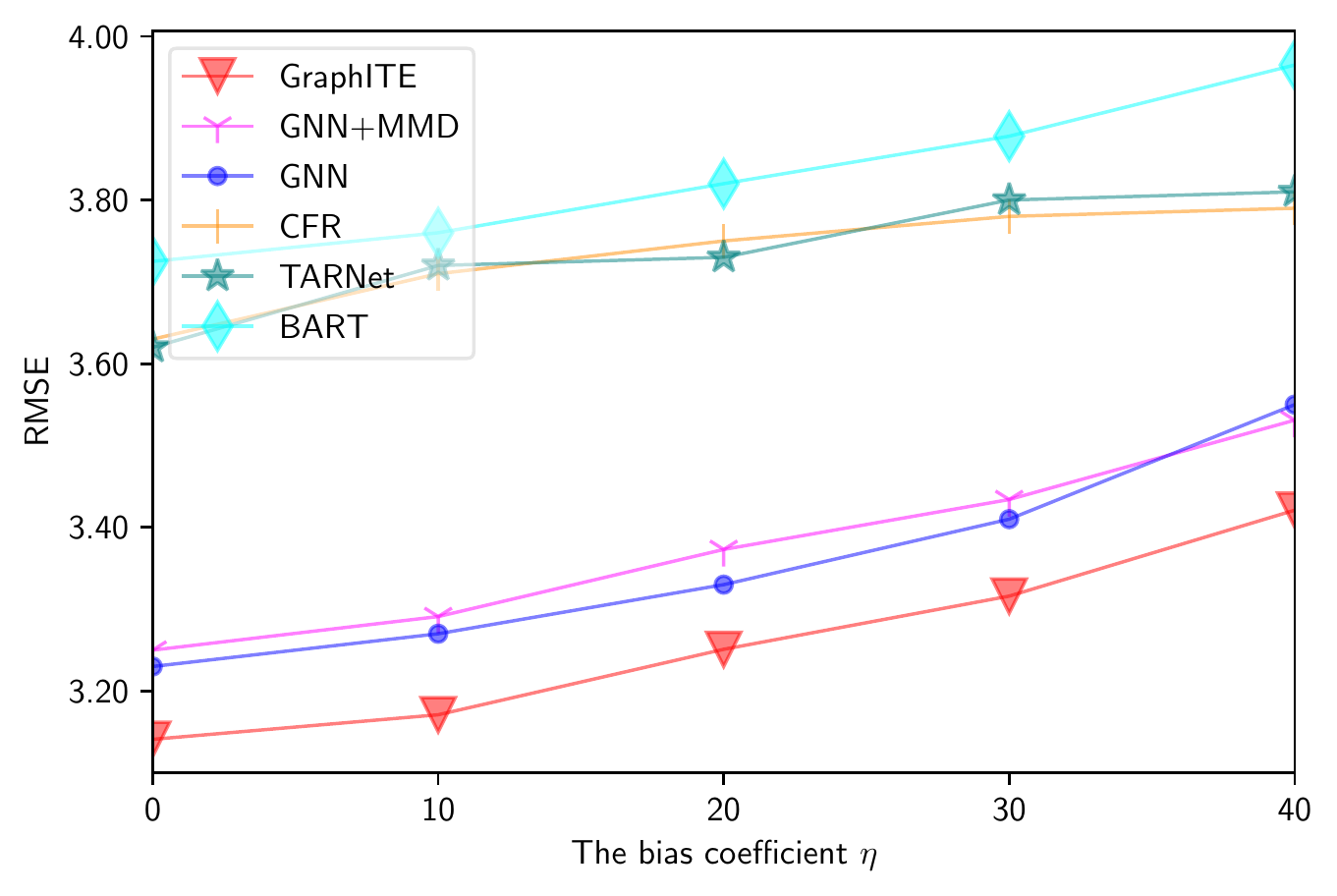}
      \centering
   (c): RMSE~(GDSC)
 \end{minipage}
 \centering
  \begin{minipage}{0.245\hsize}
   \includegraphics[width=\linewidth]{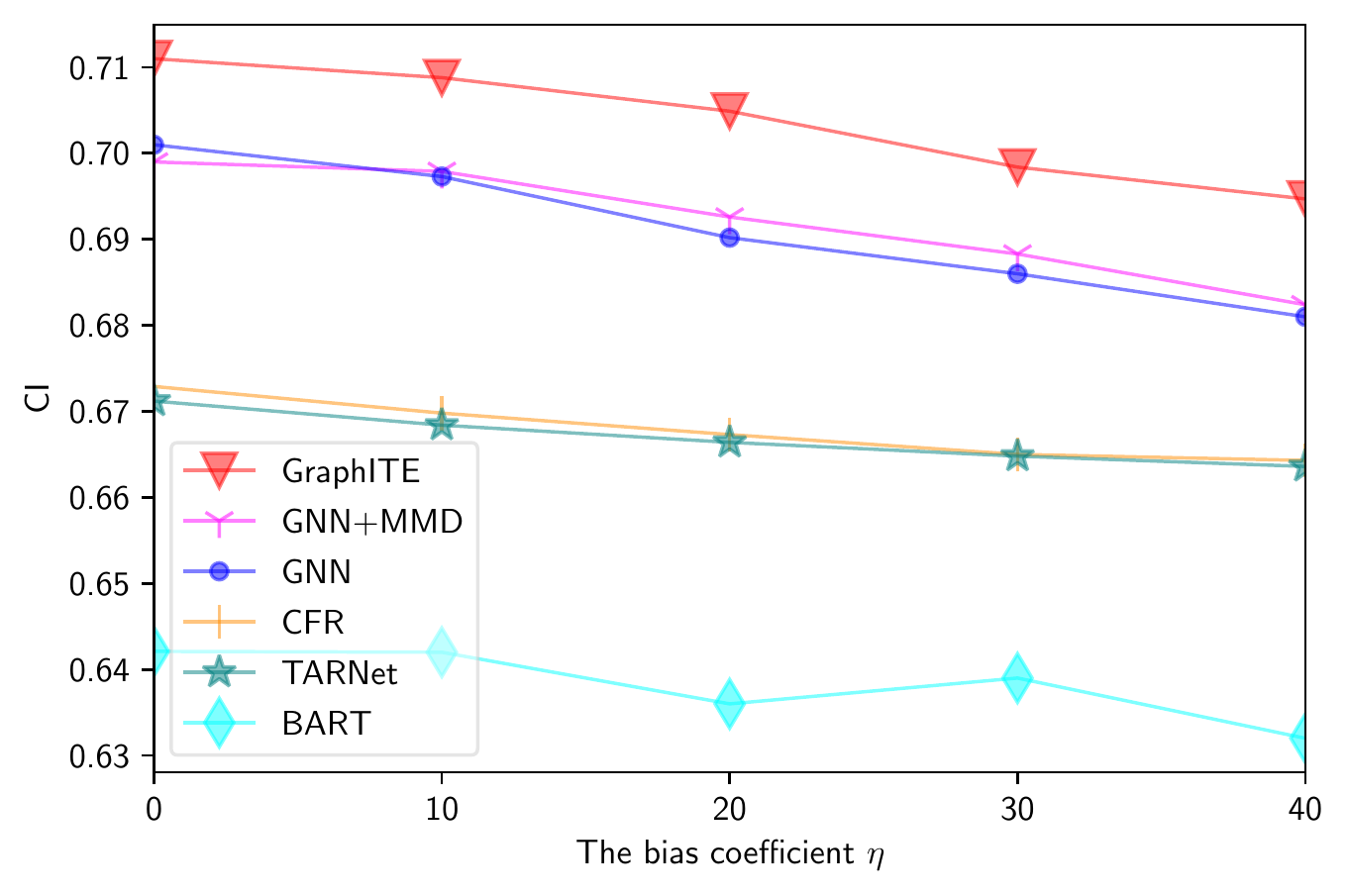}
      \centering
   (d): CI~(GDSC)
 \end{minipage}\\
 \caption{\small{Predictive performance depending on the bias coefficient $\eta$. A large $\eta$ indicates a larger selection bias. 
GraphITE shows its strong and stable tolerance to the biases and consistently performs the best in the whole range.}}\label{fig:bias}
\end{figure*}

 \begin{figure*}[tb]
\centering
\begin{minipage}{0.245\hsize}
   \includegraphics[width=\linewidth]{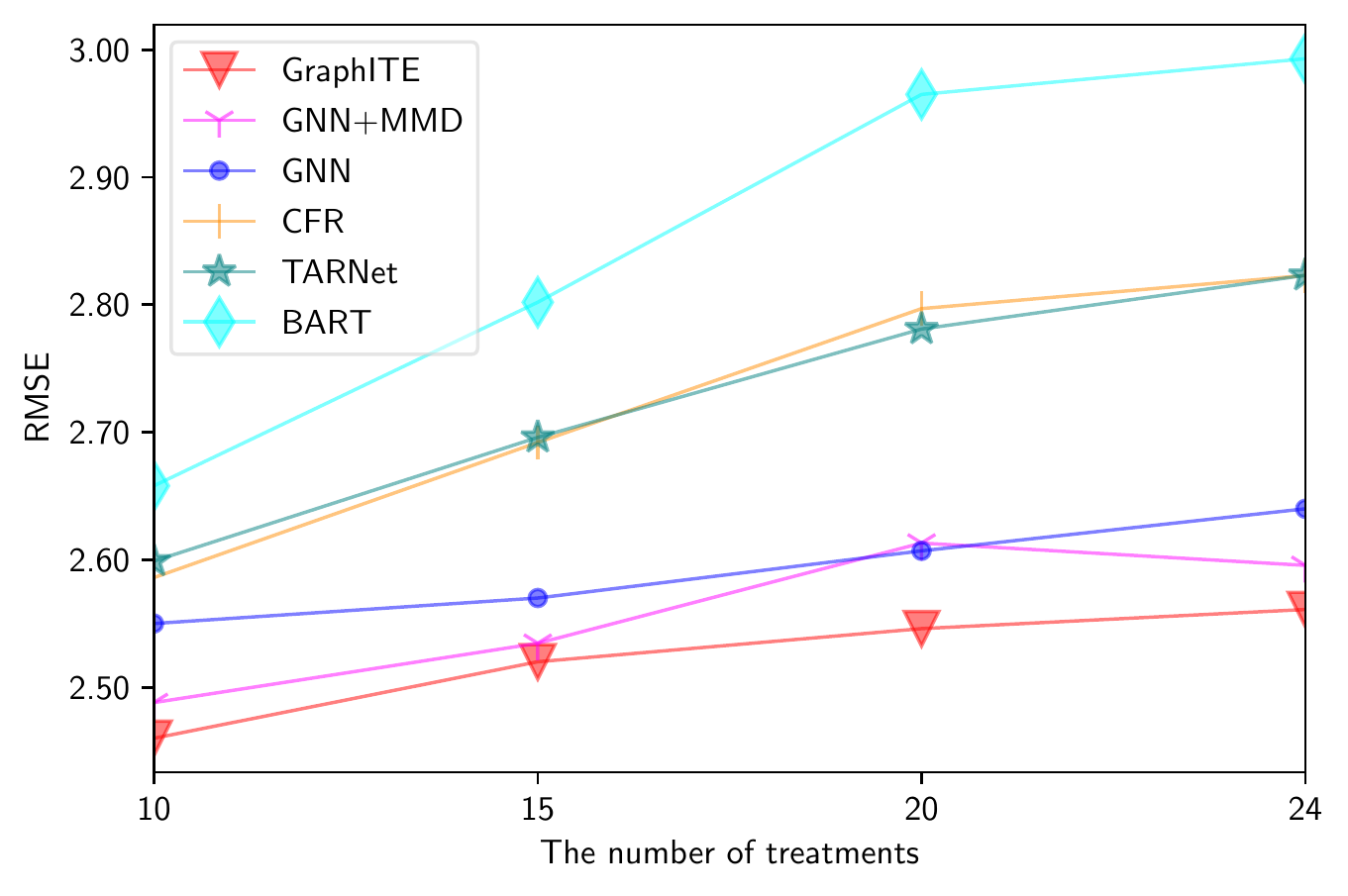}
      \centering
   (a): RMSE~(CCLE)
 \end{minipage}
 \begin{minipage}{0.245\hsize}
   \includegraphics[width=\linewidth]{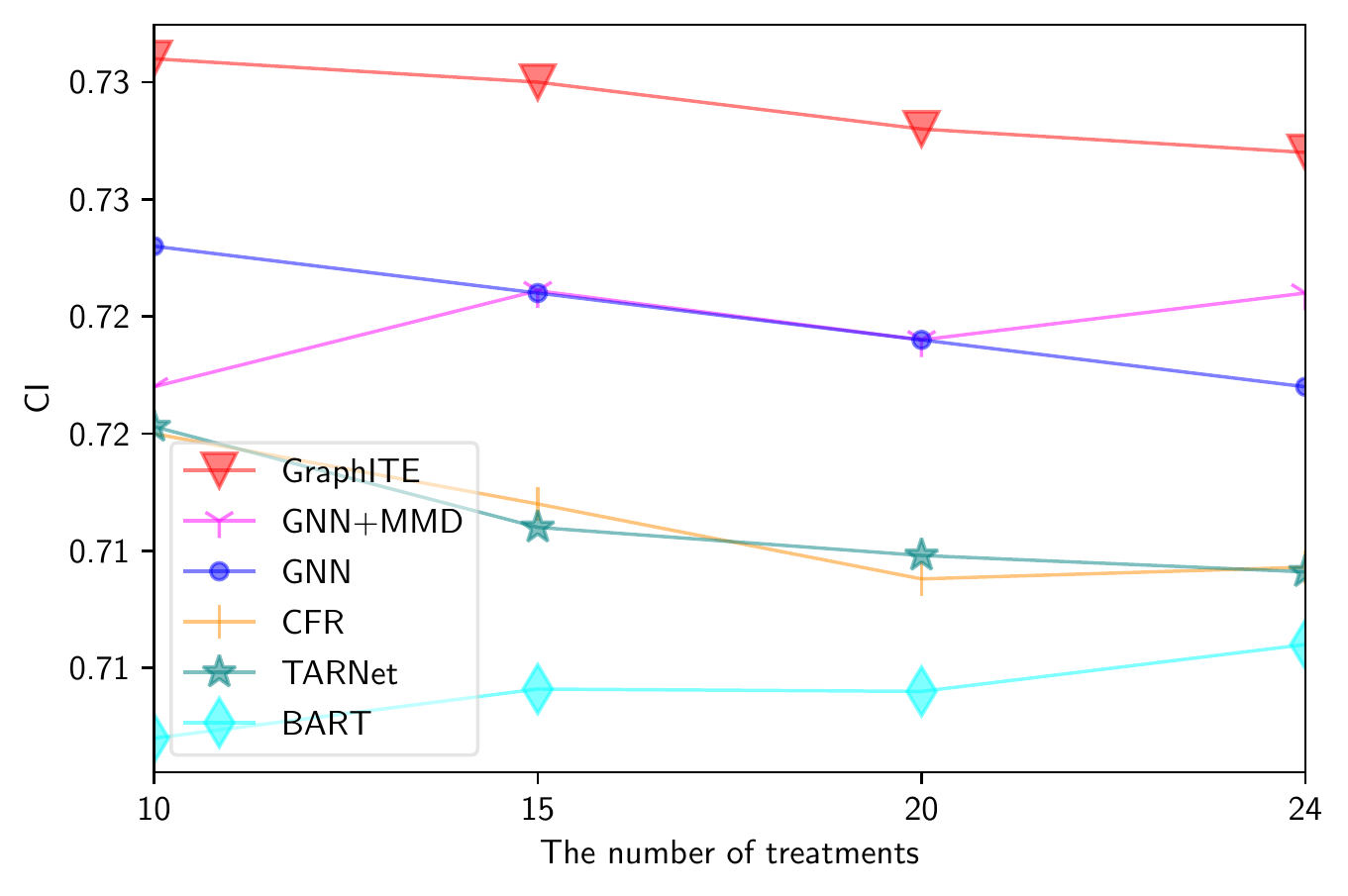}
      \centering
   (b): CI~(CCLE)
 \end{minipage}
 \begin{minipage}{0.245\hsize}
   \includegraphics[width=\linewidth]{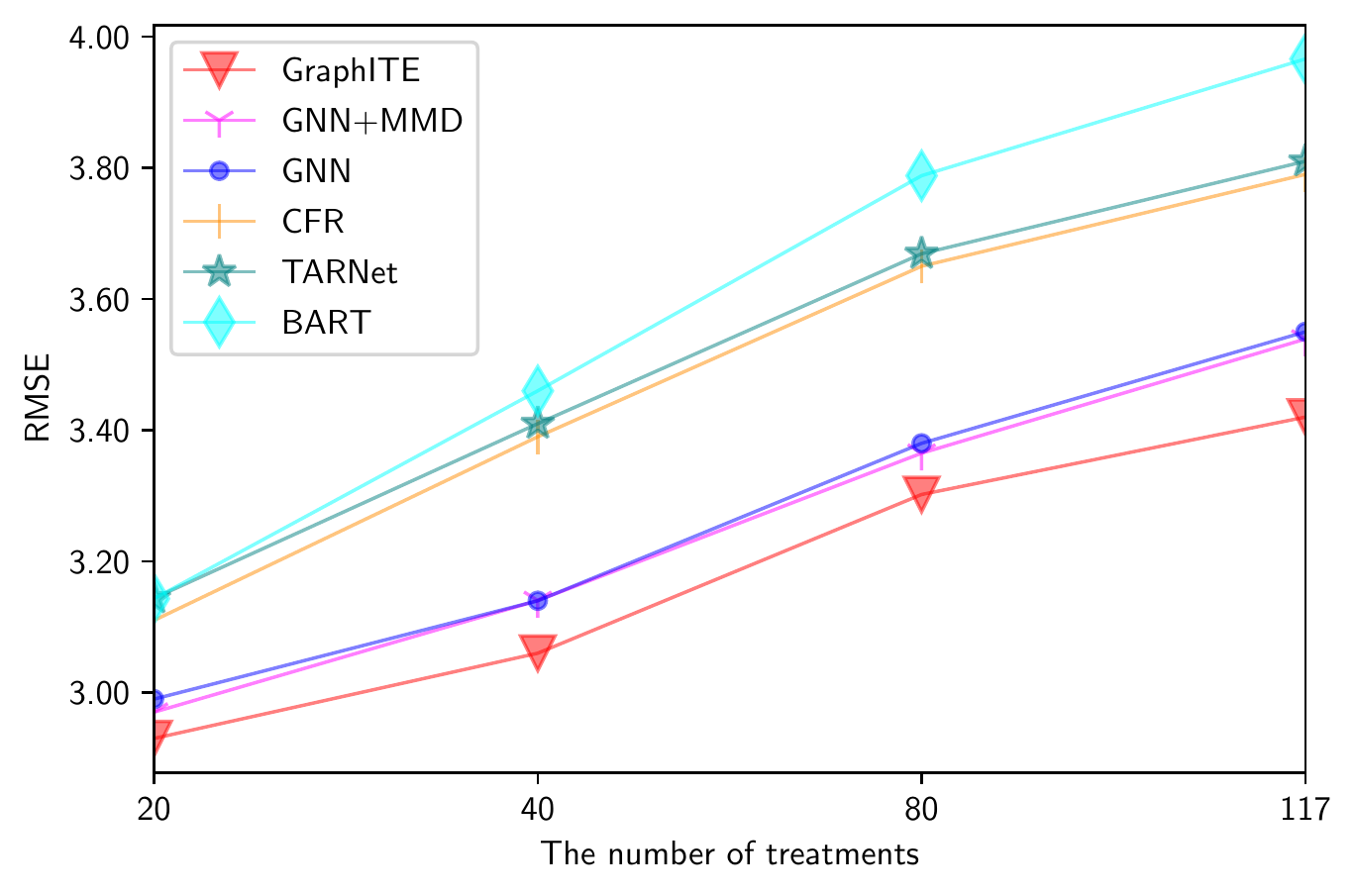}
      \centering
   (c): RMSE~(GDSC)
 \end{minipage}
 \centering
  \begin{minipage}{0.245\hsize}
   \includegraphics[width=\linewidth]{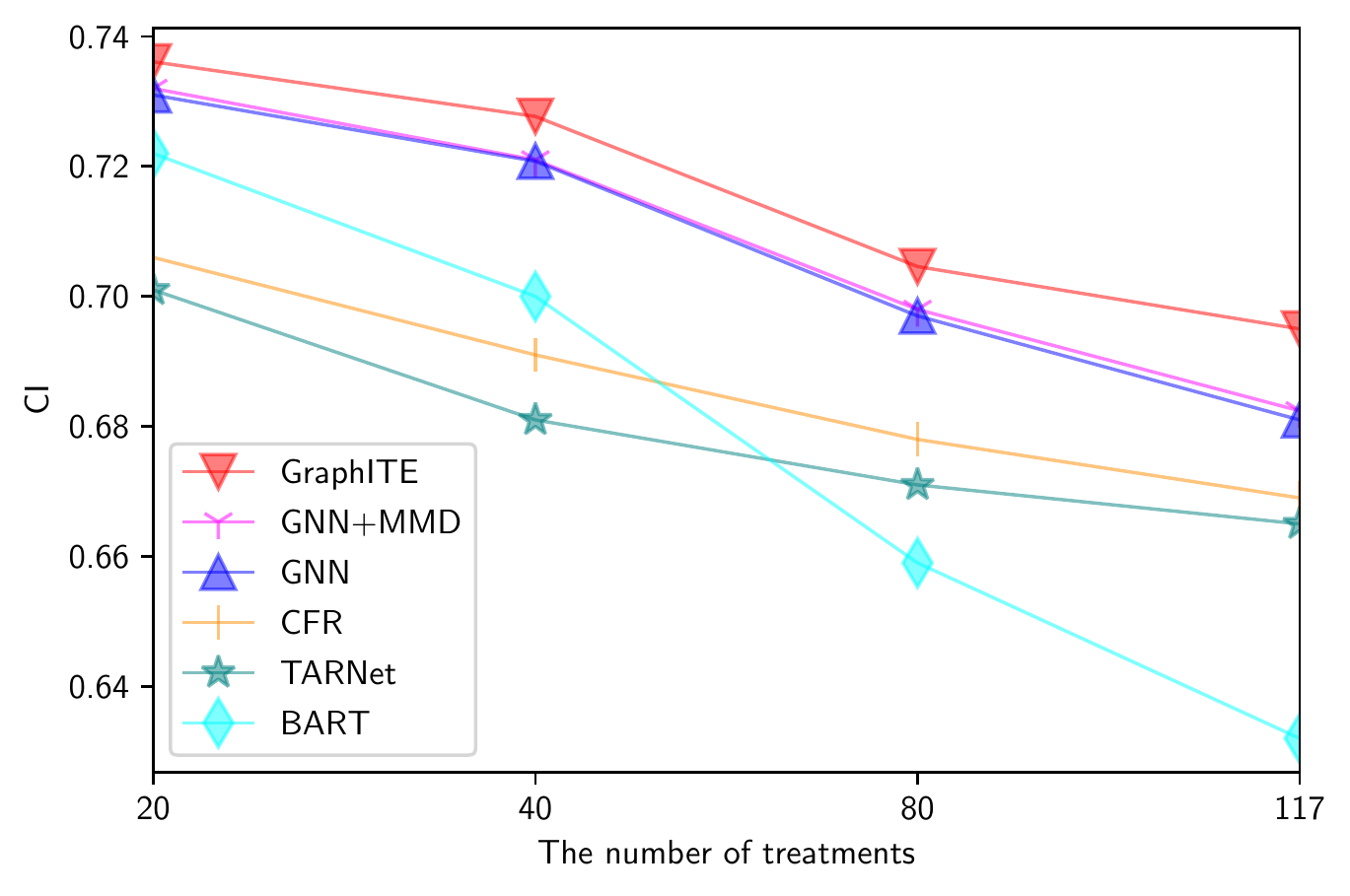}
      \centering
   (d): CI~(GDSC)
 \end{minipage}\\
 \caption{\small{Predictive performance depending on the the number of treatments. Whereas the baseline methods get degraded as the number of treatments increase, graphite shows relatively robust to its increase and achieves the best performances, especially in the larger dataset~(GDSC).
 }}\label{fig:scalability}
\end{figure*}

 \begin{figure*}[tb]
\centering
\begin{minipage}{0.245\hsize}
   \includegraphics[width=\linewidth]{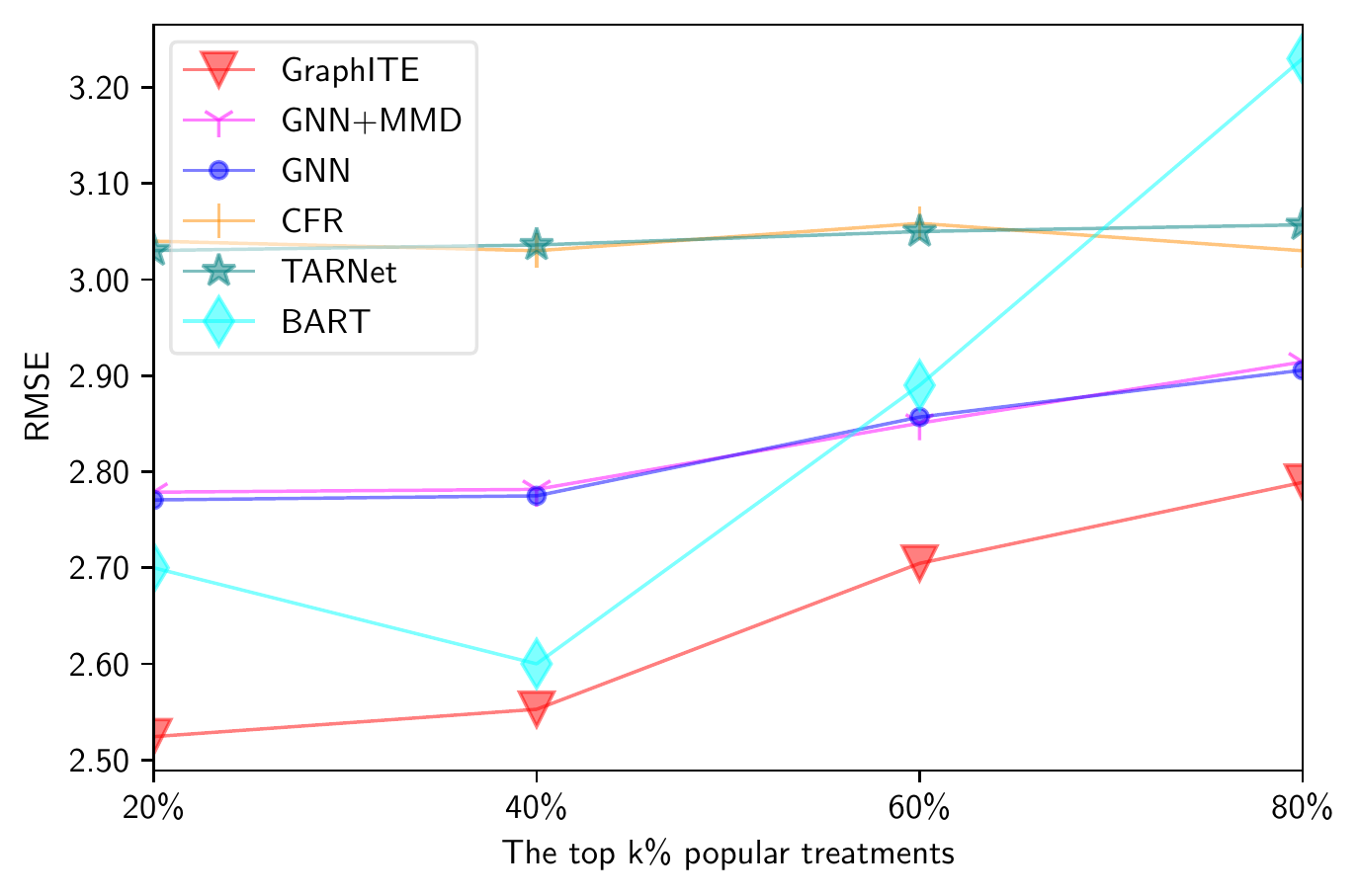}
      \centering
   (a): RMSE~(CCLE)
 \end{minipage}
 \begin{minipage}{0.245\hsize}
   \includegraphics[width=\linewidth]{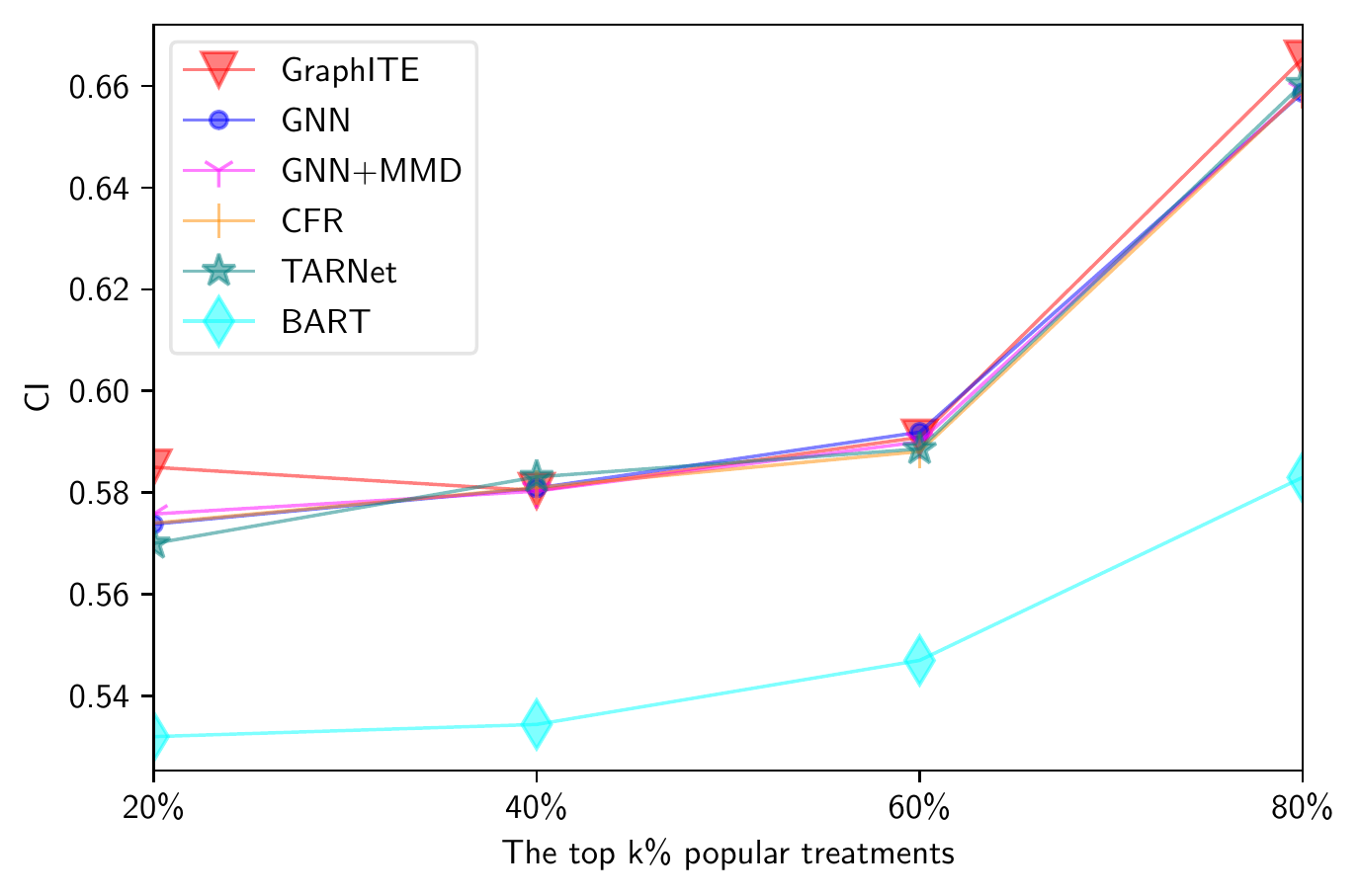}
      \centering
   (b): CI~(CCLE)
 \end{minipage}
 \begin{minipage}{0.245\hsize}
   \includegraphics[width=\linewidth]{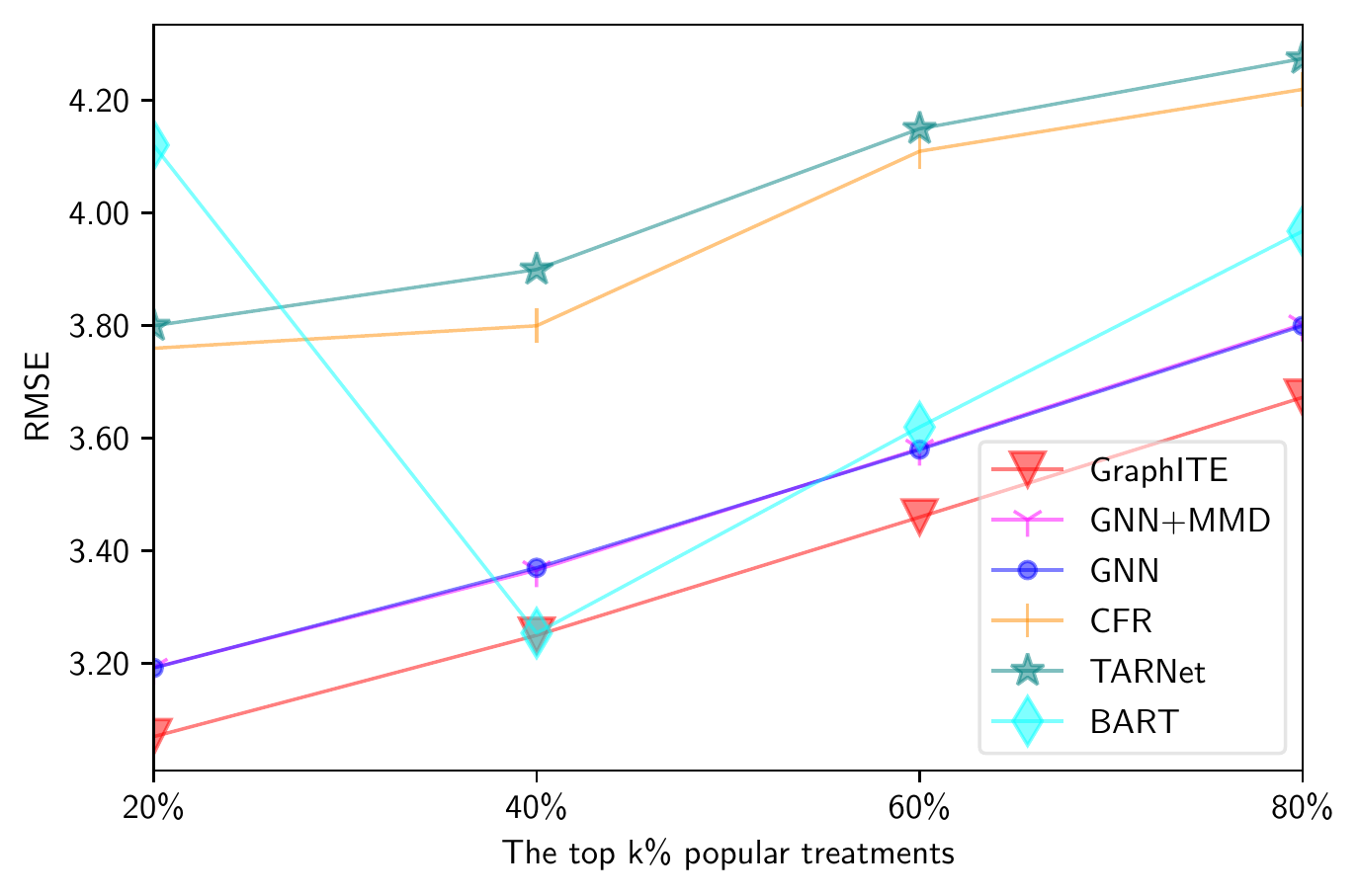}
      \centering
   (c): RMSE~(GDSC)
 \end{minipage}
 \centering
  \begin{minipage}{0.245\hsize}
   \includegraphics[width=\linewidth]{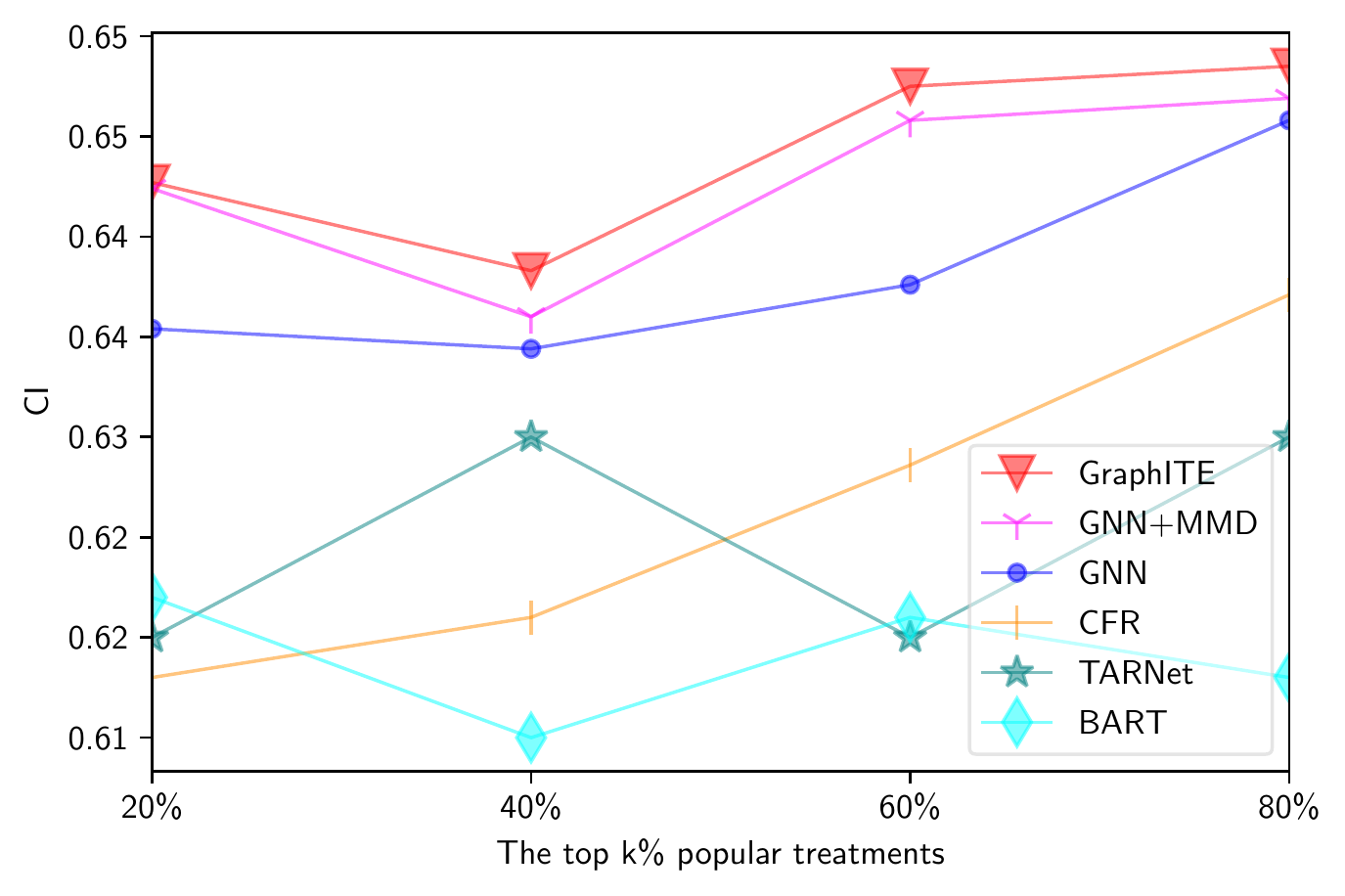}
      \centering
   (d): CI~(GDSC)
 \end{minipage}\\
 \caption{\small{Predictive performance depending on treatment popularity. From the RMSE results, the methods that do not rely on treatment graph information (CFR, TARNet, BART) suffer from a lack of data especially for unpopular treatments. From the CI results, the methods that have no bias mitigation mechanism put too much attentions on popular treatments (i.e., difficult in terms of ranking) treatments, and perform suboptimally. GraphITE shows the most stable and best performance on every group.}}\label{fig:few}
\end{figure*}

\begin{table*}[t]
    \caption{\small{Performance comparison of different methods on the CCLE and  GDSC dataset in terms of RMSE and CI. $^\dagger$ and $\ddagger$ indicate statistically significantly better performance of the proposed GraphITE than the baseline by the paired $t$-test with $p<0.05$ and $p<0.01$, respectively. The bold results indicate the statistically significant best results. The shaded rows indicate the GNN-based methods. Lower RMSEs are better, and higher CIs are  better.}}\label{table:unobserved_results}
\centering
\begin{tabular}{ccccccccc}
\toprule[2pt]
                & \multicolumn{2}{c}{ \bf{CCLE} }        &              &  \multicolumn{2}{c}{ \bf{GDSC} } \\[3pt] \cmidrule{1-6}
Method                       & RMSE & CI & \multicolumn{1}{l}{} &  RMSE &  CI  \\[3pt] \cmidrule{1-6}
\multicolumn{1}{c}{Mean }    &        $3.458_{\pm 1.301}$       &       $-$         & &  $^\dagger 4.705_{\pm 0.702}$   & $-$  \\[2pt]
\rowcolor{Gray}
\rowcolor{Gray}
 GNN  &        $3.920_{\pm{0.932}}$       &       $0.551_{\pm{0.130}}$        &&         $^\ddagger4.646_{\pm{0.631}}$        &       $0.570_{\pm{0.061}}$        \\[2pt]
\rowcolor{Gray}
 GNN+MMD  &        $3.903_{\pm{0.923}}$       &       $0.549_{\pm{0.132}}$        & &         $^\ddagger4.640_{\pm{0.674}}$        &       $0.574_{\pm{0.061}}$        \\[2pt]
\rowcolor{Gray}
 GraphITE  &        $ 3.637_{\pm{0.905}}$       &       $ 0.545_{\pm{0.114}}$        &   &      $\bf 4.482_{\pm{0.595}}$        &       $ 0.569_{\pm{0.054}}$        \\
\bottomrule[2pt]
    \end{tabular}
        \end{table*}

\subsection{Baseline methods}

We compare GraphITE with the following six baselines.
(i) Ordinary least squares linear regression~(OLS) concatenates two vectors, the covariate vector and treatment vector coded as a one-hot vector, which is used as the input.  
(ii) Bayesian additive regression trees~(BART)~\cite{chipman2010bart,hill2011bayesian} predicts the outcomes by an ensemble of multiple regression trees; we used a Python implementation of BART\footnote{\url{https://github.com/JakeColtman/bartpy}}.
(iii) Treatment embedding method exploits low-dimensional representations of treatments to deal with a large number of treatments. Each treatment is associated with a low-dimensional vector, which is input to a neural network, as well as a covariate vector.
Note that this method does not use the graph structures of the treatments at all.
(iv) TARNet~\cite{shalit2017estimating} is a deep neural network model with shared layers for representation learning and different layers for outcome prediction for treatment and control instances. 
(v) Counterfactual regression~(CFR)~\cite{shalit2017estimating} is one of the state-of-the-art deep neural network models based on balanced representations between treatment and control instances; we used the MMD as its IPM.
Following previous studies~\cite{yoon2018ganite,saini2019multiple}, we extend the CFR~\cite{shalit2017estimating} to the multiple-treatment setting; we regard the most frequent treatment as the control treatment.
(vi) GANITE~\cite{yoon2018ganite} is another state-of-the-art deep neural network model based on GAN.
It trains a TARNet-like generator that generates counterfactual outcomes, and a discriminator tells whether outcomes come from the generator or the real distribution. 
In the original GANITE, the discriminator just tries to solve a binary classification; on the other hand in our setting, GANITE has to solve multi-class classification to tell which outcome is the genuine one.

In addition to the use of graph structured treatments, one of the key features of GraphITE is the bias mitigation using HSIC regularization; therefore, we use the versions without it our baseline methods for the ablation study. 
We also tested several variants of GraphITE:
(vii) a variant with no bias mitigation that does not have the HSIC regularization term and only uses a GNN, which we refer to as ``GNN'' hereafter and
(viii) another variant using MMD regularization instead of HSIC regularization. 
We used the same approach as CFR to deal with multiple treatments.
We denote it by ``GNN+MMD".

\subsection{Experimental setting}
As the evaluation metrics, we employ the root mean square error~(RMSE)~of all target--treatment pairs in the test set defined as:
\begin{equation}
\text{RMSE}=\sqrt{\frac{1}{N^{\text{test}}}\frac{1}{\mid\mathcal{T}\mid}\sum_{i=1}^{N^\text{test}}\sum_{t=1}^{\mid\mathcal{T}\mid}(y^{t}_{i}-f(x_i,t)     )^{2}},
\end{equation}
where  $N^{\text{test}}$ is the number of target individuals included in the test dataset.
We also employ the concordance index (CI)~\cite{harrell1996multivariable}  to evaluate predictive performance in terms of ranking accuracy, which has been widely used in previous studies~\cite{kurilov2020assessment,safikhani2017gene}. 
The CI is defined as
\begin{equation}\label{eq:defCI}
\text{CI}=\frac{1}{N^\text{test}} \sum_{i=1}^{N^\text{test}} \sum_{t,u \mid y_i^t>y_i^u}  \frac{\theta(f(x_i,t)  - f(x_i,u) )}{|\{t,u \mid y_i^t>y_i^u\}| },
\end{equation}
where $\gamma$ is the Heaviside step function defined as 
\begin{equation}
\theta(x)=
\begin{cases}
    1, & x > 0 \\
    0.5,        & x=0\\
    0 & x<0
\end{cases}.
\end{equation}
Note that the CI is identical to ROC-AUC when all outcomes are binary.

We split the whole individuals into $80\%$, $10\%$, and $10\%$ for training, validation, and testing sets, respectively.
We report the average results of $50$ different trials. 
Note that while we sample the factual treatments in the training and validation sets following the biased sampling scenario explained in the Dataset section, all of the treatments are included in the test set because we want our prediction model to perform uniformly well on all treatments.

In GraphITE, to promote effective feature extraction from small data, we pre-train the GNN $\psi$ on regression tasks using three popular molecular datasets: ESOL, FreeSolv, and  Lipophilicity, provided by MoleculeNet~\cite{wu2018moleculenet}\footnote{\url{http://moleculenet.ai/}}. To the best of our observation, GraphITE performs slightly better with pre-training than the one without pre-training. We believe that this phenomenon may be caused by the small size of the training data, and is the one of limitations of this study to be addressed in future work.
For the HSIC regularization, we use the normalized version of the HSIC~(nHSIC), defined as  
\begin{equation}
\text{nHSIC}(\phi, \psi) =\frac{\text{tr}({\bf K}^{\Phi}{\bf H}{\bf K}^{\Psi}{\bf H})}{ \| {\bf K}^{\Phi}{\bf H} \|\| {\bf K}^{\Psi}{\bf H} \|},
\end{equation}
where $\| \cdot\|$ denotes the Frobenius norm.
The regularization parameter $\lambda$ is optimized in $\{10^{-3}, 10^{-2},\ldots, 10^3\}$ based on the RMSE for the validation set. 
We set the number of representation dimensions for target individuals and graph-structured treatments to $64$ because we did not observe the significant differences in $\{16,32,64,128\}$ in the both datasets.
Similarly, we set the numbers of layers of $\phi, \psi$, and $g$ to $3$.

\subsection{Results}

Table~\ref{table:results} summarizes the predictive performances of the different methods for $\eta=40$ (i.e., the strongest bias). 
In the remainder, we report the experimental results when we set $\eta=40$ unless otherwise stated.

The deep learning-based methods (Treatment Embedding, TARNet, CFR) outperform the na\"ive methods, such as Mean and OLS.
BART also gives the comparable performance. 
However, we observe GANITE performs poorly in comparison with the other deep learning-based methods. We believe this is because of the difficulty in learning the GAN models, that is, GANITE has to solve many-class classification with a limited amount of training data. 
The existing bias mitigation methods that are simply extended to multiple treatments~(CFR and GNN+MMD)~do not not show significant improvements over the corresponding original ones~(TARNet and GNN). 
By contrast, GraphITE achieves the best performance, which is statistically significant against all the baselines, and it demonstrates its effectiveness, especially on the larger dataset~(GDSC).
The merit of exploiting the graph structures associated with treatments can also be seen in Table~\ref{table:results}.
The GNN-based methods~(shaded rows)~perform better than  the methods that neglect graph-structured information.

Figure~\ref{fig:reg} shows the sensitivity of the results to the strength of HSIC regularization; although the optimal choices significantly improves the performance, none of the other choices harm the performance.
The results also highlight the effectiveness of the HSIC as the choice for the regularization term; MMD shows no remarkable improvement over the plain GNN because it cannot handle many treatments efficiently, whereas GraphITE using HSIC regularization shows distinct improvements.

Now, we investigate the robustness of GraphITE against selection bias. 
Figure~\ref{fig:bias} shows the performances for different bias strengths, where a larger $\eta$ represents a larger selection bias. 
GraphITE shows its strong and stable tolerance to the biases and consistently performs best under all  bias strength settings.

The impacts of the number of treatments are shown in Figure~\ref{fig:scalability}. 
For small numbers of treatments, both GraphITE and the other baseline methods perform similarly well; 
however, the baseline methods, especially BART, degrade the performances as the number increase. 
GNN+MMD does not show improvements from the plain GNN, particularly on the larger dataset~(GDSC). 
GraphITE shows the remarkable robustness to the selection bias, even with large numbers of treatments.

Next, we investigate the predictive performance based on treatment popularity, as shown in Figure~\ref{fig:few}. 
We focus on the treatment groups that are grouped by their popularity, namely, the top 20\%, 20--40\%, 40--60\%, 60--80\% most popular treatment groups. 
As can be seen from the RMSE results, the methods that do not rely on treatment graph information (CFR, TARNet, and BART) suffer from a lack of training data especially for unpopular treatments; On the other hand, the methods exploiting the auxiliary information mitigate this problem. 

Now we turn to the CI results. From its definition (\ref{eq:defCI}), CI measures ranking performance. It is more difficult to estimate accurate ranking for popular treatments rather than less popular treatments, because in our bias setting, more popular treatments have larger outcomes, and they are more heterogeneous sets than less popular ones.
On the other hand, it is rather easy to obtain small RMSEs for popular treatments because their outcomes have small variances.
The methods that have no bias mitigation mechanism put too much attentions on such difficult treatments (in terms of ranking), and eventually perform suboptimally, while GraphITE shows the most stable and best performance on every group.

Finally, we investigate the predictive performance on the unobserved ``zero-shot" treatments that are not included in training data. 
We keep $30\%$ of the entire treatments aside in advance as the validation and target unobserved treatments.  
Table~\ref{table:unobserved_results} shows the prediction accuracy for the unobserved treatments when we set $\eta=40$. 
Note that the existing methods such as CFR are incapable of dealing with this setting.
In the smaller dataset, CCLE, the  selection bias prevents the models from working appropriately, and all of the variants perform worse than the simply baseline taking the mean of training data in terms of RMSE. 
On the other hand in the larger dataset, GDSC, GraphITE achieves the best RMSE, while the other methods still suffer from the bias. 
However, we do not observe significant improvements in terms of CI in the both datasets.

\section{Conclusion}
In this study, we proposed GraphITE, which can handle graph-structured treatments in order to achieve better treatment effect estimation even when the number of treatments is large.
GraphITE is based on the recent developments of deep neural networks and representation learning, namely, GNNs and HSIC regularization, which contribute to improving estimation accuracy of complex -structured treatments from biased observational data. 
In addition, GraphITE is applicable to previously unobserved ``zero-shot" treatments, which the existing ITE estimation methods are intrinsically not capable of dealing with.
In the experiments on two real-world drug response datasets, GraphITE achieved the best performances in terms of RMSE and CI when compared to the various baselines. 
In particular, we observed a significant improvement when the effect of selection bias and the number of treatments were large.
A potential future direction is to consider other types of complex structured data, such as texts, images, and videos. 
We also plan to apply GraphITE to much larger datasets, in which we expect further  improvements.

\section*{Acknowledgments}
This work was supported by JSPS KAKENHI Grant Number 21J14882 and 20H04244.
\bibliographystyle{ACM-Reference-Format}
\balance
\bibliography{MAIN}

\end{document}